\documentclass[journal]{IEEEtran}
\usepackage{float}

\usepackage[utf8]{inputenc}

\usepackage{tabularx}
\usepackage{subcaption}

\usepackage{booktabs}

\usepackage{multirow}
\usepackage{hyperref}
\usepackage{threeparttable}
\usepackage[most]{tcolorbox}

  \usepackage{graphicx}

\usepackage{todonotes}

\newcommand{\reg}[1]{\textsuperscript{\textregistered}}

\begin{document}
%

\title{Methods and open-source toolkit for \\analyzing and visualizing challenge results}

%
%
%

\author{Manuel Wiesenfarth, Annika Reinke, Bennett A. Landman, Manuel Jorge Cardoso, Lena Maier-Hein*, Annette Kopp-Schneider*
\thanks{M. Wiesenfarth and A. Kopp-Schneider are affiliated with the Division of Biostatistics, German Cancer Research Center (DKFZ), 69120, Heidelberg, Germany (e-mail: m.wiesenfarth@dkfz-heidelberg.de; kopp@dkfz-heidelberg.de).}
\thanks{A. Reinke and L. Maier-Hein are affiliated with the Division Computer Assisted Medical Interventions (CAMI), German Cancer Research Center (DKFZ), 69120, Heidelberg, Germany (e-mail: l.maier-hein@dkfz-heidelberg.de; a.reinke@dkfz-heidelberg.de).}
\thanks{B. A. Landman is affiliated with the Electrical Engineering, Vanderbilt University, Nashville, TN, 37235-1679, USA (e-mail: bennett.landman@vanderbilt.edu).}
\thanks{M. J. Cardoso is affiliated with the School of Biomedical Engineering \& Imaging Sciences, King’s College London, London, WC2R 2LS, UK (e-mail: m.jorge.cardoso@kcl.ac.uk).}
\thanks{*: shared senior authors}
}

\newlength{\figurewidth}
\setlength{\figurewidth}{0.4\textwidth}

\graphicspath{{./figuresShort/}}

\maketitle

\begin{abstract}
Biomedical challenges have become the de facto standard for benchmarking biomedical image analysis algorithms. While the number of challenges is steadily increasing, surprisingly little effort has been invested in ensuring high quality design, execution and reporting for these international competitions. Specifically, results analysis and visualization in the event of uncertainties have been given almost no attention in the literature. Given these shortcomings, the contribution of this paper is two-fold: (1) We present a set of methods to comprehensively analyze and visualize the results of single-task and multi-task challenges and apply them to a number of simulated and real-life challenges to demonstrate their specific strengths and weaknesses; (2) We release the open-source framework \textit{challengeR} as part of this work to enable fast and wide adoption of the methodology proposed in this paper. Our approach offers an intuitive way to gain important insights into the relative and absolute performance of algorithms, which cannot be revealed by commonly applied visualization techniques. 
This is demonstrated by the experiments performed within this work. Our framework could thus become an important tool for analyzing and visualizing challenge results in the field of biomedical image analysis and beyond.
\end{abstract}


%
\IEEEpeerreviewmaketitle


\section{Introduction}
\label{sec:introduction}
\IEEEPARstart{I}n the last couple of years, \textit{grand challenges} have evolved as the standard to validate biomedical image analysis methods in a comparative manner~\cite{maierhein2018rankings}. The results of these international competitions are commonly published in prestigious journals, and challenge winners are sometimes awarded with huge amounts of prize money. Today, the performance of algorithms on challenge data is essential, not only for the acceptance of a paper, but also for the individuals' scientific careers and to give the algorithms the opportunity to be used in a clinical setting. Given the scientific impact of challenges, it is surprising that there is a huge discrepancy between their impact and quality control. Challenge reporting is usually poor, the design across challenges lacks common standards and challenge rankings are sensitive to a range of challenge design parameters. This has all been demonstrated by a study on biomedical image analysis competitions \cite{maierhein2018rankings}.

\begin{figure}
\begin{tcolorbox}[title= Dot- and boxplot]
\centering
\begin{subfigure}{\figurewidth}
    \centering
    \caption{c\_ideal~~~~~~~~~~~~~~~~~(b) c\_random}\vspace{-.5em}
    \label{subfig:synthetic:raw:boxplot:a}
    \includegraphics[width=.9\figurewidth]{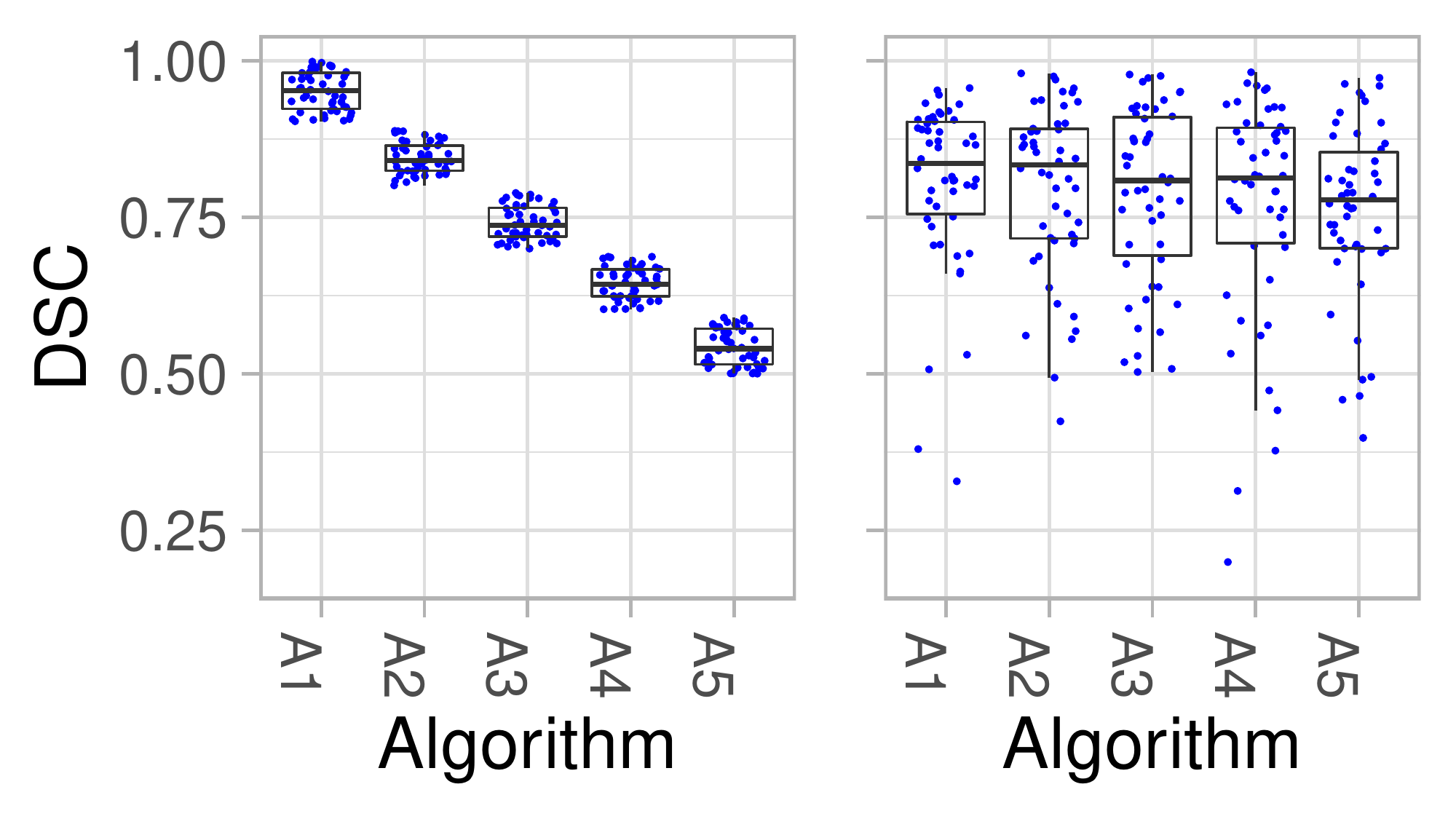} 
\end{subfigure}
\caption{\textit{Dot- and boxplots} for visualizing the assessment data separately for each algorithm. Boxplots representing descriptive statistics for all test cases (median, quartiles and outliers) are combined with horizontally jittered dots representing individual test cases.}\label{fig:synthetic:raw:boxplot}
\end{tcolorbox}
\end{figure}

As rankings are the key to identifying the challenge winner, this last point is crucial, yet most publications of challenges ignore it. Instead, the presentation of results in publications is commonly limited to tables and simple visualization of the metric values for each algorithm. 
In fact, from all the challenges that were analyzed in ~\cite{maierhein2018rankings} and had their results published in journals ($n = 83$), 27\% did not use any visualization method apart from a table listing the final rank for each method. 39\% of challenges included boxplots to visualize challenge results, whereas not a single challenge visualized uncertainties. This is critical because crucial information on the stability of the ranking is not conveyed. Consider for example the two example challenges \textit{c\_random} and \textit{c\_ideal} depicted in Fig. \ref{fig:synthetic:raw:boxplot}. 
The rankings of these challenges are identical, although the distributions of metric values are radically different: For the challenge \textit{c\_random}, there should in fact be only one shared rank for all algorithms, because the metric values for the different methods  were   drawn from the same distribution (for details see sec. \ref{sec:methods:example:failureRandom}). In contrast, the first ranked algorithm of challenge \textit{c\_ideal} is the clear winner. 

Overall, our study of past challenges revealed that advanced visualization schemes (beyond boxplots and other basic methods) for providing deeper insights into the performance of the algorithms were not applied in any of the papers. A possible  explanation is the lack of standards for challenge data analysis and visualization. While the topic of visualization
is essential in the field of biomedical data analysis in general~\cite{o2018visualization}, we are not aware of any prior work in the field of challenge data analysis.  
%


The purpose of this paper is therefore  to propose methodology along with an open-source framework for systematically analyzing and visualizing results of challenges. Our work will help challenge organizers and participants gain further insights into both the algorithms' performance and the assessment data set itself in an intuitive manner.  We present visualization approaches for both challenges designed around a single task (\textit{single-task challenges}) 
and for challenges comprising  multiple tasks (\textit{multi-task challenges}), such as the Medical Segmentation Decathlon (MSD)~\cite{decathlon}. 

The paper is organized as follows:
Sec.~\ref{sec:dataandpreprocessing} presents the data used for the illustration and validation of our methodology along with the data analysis methods that serve as prerequisite for the challenge visualization methods. Secs.~\ref{sec:synthetic} and~\ref{sec:synthetic:multiple} then present visualization methods for single-task and multi-task challenges, respectively, addressing the stability (effect of data variability) and robustness (effect of ranking method choice) of the challenge results. 
Sec.\ref{sec:toolkit} introduces the open source framework in which we implemented the 
methodology. 
Finally, we close with a discussion of our findings in sec.~\ref{sec:conclusion}.

\section{Data and Data Processing}
\label{sec:dataandpreprocessing}

Computing a challenge ranking is typically done using the following elements:
\begin{itemize}
    \item The challenge metric(s) used to compute the performance of a participating algorithm for a specific \textit{test case}, where a  test case encompasses all  data (including the reference annotation) that is processed to produce one result,
    \item The $m$ challenge task(s),
    \item The $p$ competing algorithms,
    \item The $n_k, k=1,\ldots,m$, test cases for each task and
    \item A rule on how to deal with missing values that occur if an algorithm does not deliver a metric value for a test case. Typically the value is set to an unfavorable value, e.g., 0 for a non-negative metric in which larger values indicate better performance.
\end{itemize}
Note that we use the term 'assessment data' in the following 
to refer to the challenge results and not to the  (imaging) data given to challenge participants. Further, we will use the term 'metric' as an equivalent to performance measure and thus is not related to the mathematical definition. 

The further course of this section introduces the data used for this paper (sec. \ref{sec:assessmentdata}) along with the basic methodology used for generating (sec. \ref{sec:methods:ranking}) and comparing (sec. \ref{sec:methods:comparison}) rankings and for computing ranking stability (sec. \ref{sec:methods:stability}).

\subsection{Assessment data}
\label{sec:assessmentdata}
We use three assessment data sets 
corresponding to three different (simulated and real) challenges for this manuscript: two simulated challenges (\textit{c\_ideal} 
and \textit{c\_random}) to illustrate the analysis and visualization methodology and one real challenge, \textit{c\_real}, to apply our method to a complex real-world example. 
\subsubsection{c\_ideal: Best-case scenario with 
ideal assessment data}\label{sec:methods:example:ideal}
We generated synthetic assessment data in which the ranking of the five algorithms $A_1$ to $A_5$ is clear and indisputable. Data mimic Dice Similarity Coefficient (DSC) \cite{dice1945measures} measurements which are often used within medical image segmentation tasks to assess the overlap between two objects and which generate values between 0 (worst) and 1 (best). We simulated $n=50$ uniform samples (representing challenge test cases) from $[0.9,1),[0.8,0.9),[0.7,0.8),[0.6,0.7)$ and $[0.5,0.6)$ for algorithms $A_1,A_2,\ldots,A_5$, respectively.


\subsubsection{c\_random: Fully random scenario where differences are due to chance}\label{sec:methods:example:failureRandom}
250 random normal values with a mean of 1.5 and variance 1 were drawn and transformed by the logistic function to obtain a skewed distribution on $[0;1]$. These were then assigned to algorithms $A_1$ to $A_5$, resulting in $n=50$ test cases. Thus, there is no systematic difference between the algorithms, any difference can be attributed to chance alone.

\subsubsection{c\_real: Real-world assessment data example}\label{sec:methods:example:real}
We apply the visualization methods to a real-world example, using challenge results from the MSD challenge \cite{decathlon}, organized within the scope of the Conference on Medical Image Computing and Computer Assisted Interventions (MICCAI) 2018. The challenge specifically assesses generalization capabilities of algorithms and comprises  ten different 3D segmentation tasks on ten different anatomical structures (17 sub-tasks due to multiple labels in some of the data sets).
For illustration purposes, we selected 9 of the 17 (sub-)tasks, all from the training phase of the MSD, labeled $T_1$ to $T_9$. Our analysis was executed using all participating algorithms $A_1$ to $A_{19}$ and the DSC as performance measure. 
Since the aim of the present paper is to exemplify visualization methods and not to show performance of algorithms, the challenge results were pseudonymized. For algorithms not providing a DSC value for a certain test case, this missing metric value was set to zero.

\subsection{Ranking methods}\label{sec:methods:ranking}
Many challenges produce rankings of the participating teams, often separately for multiple tasks. In general, several strategies can be used to obtain a ranking, but these may lead to different orderings of algorithms and thus different winners.
The most prevalent approaches are: 
\begin{itemize}
\item \textit{Aggregate-then-rank:} The most commonly applied method begins by aggregating metric values across all test cases (e.g., with the mean, median or another quantile) for each algorithm. This aggregate is then used to compute a rank for each algorithm. 
\item \textit{Rank-then-aggregate:} 
Another method begins, conversely, with computing a rank for each test case for each algorithm ("rank first"). The final rank is based on the aggregated test-case ranks. Distance-based approaches for rank aggregation can also be used (see sec. \ref{sec:methods:comparison}). 
\item \textit{Test-based procedures:} In a complementary approach, statistical hypothesis tests are computed for each possible pair of algorithms to assess differences in metric values between the algorithms. Then ranking is performed according to the resulting relations (e.g., \cite{demvsar2006statistical}) or 
according to the number of significant one-sided test results (e.g. for illustration, see Supplementary Discussion in \cite{maierhein2018rankings}). In the latter case, if algorithms have the same number of significant test results then they obtain the same rank. Various test statistics can be used. 
\end{itemize}
When a ranking is given, ties may occur, and a rule is required to dictate how to manage them. In the context of challenges, the rank for tied values is assigned the minimum of the ranks. For example, if the two best algorithms get the same rank, they are both declared winners. %
Generally, the larger the number of algorithms is, the greater the instability of rankings for all ranking methods and the more often ties occur in test-based procedures.

\subsection{Comparison and aggregation of rankings}\label{sec:methods:comparison}
\subsubsection{Comparison of rankings}
If several rankings are available for the same set of algorithms, the rankings can be compared using distance or correlation measures, see e.g. \cite{langville2012}. For a pairwise comparison of ranking lists, \textit{Kendall’s $\tau$} \cite{kendall1938new} is a scaled index determining the correlation between the lists. It is computed by evaluating the number of pairwise concordances and discordances between ranking lists and produces values between $-1$ (for inverted order) and $1$ (for identical order). \textit{Spearman's footrule} is a distance measure that sums up the absolute differences between the ranks of the two lists, taking the value $0$ for complete concordance and increasing values for larger discrepancies. \textit{Spearman's distance}, in turn, sums up the squared differences between the ranks of the two lists \cite{qian2019}, which in this context is closely related to the Euclidean distance. 


\subsubsection{Consensus Rankings} 
If the challenge consists of several tasks, an aggregated ranking across tasks may be desired. 
General approaches for derivation of a consensus ranking (rank aggregation) are available (see, e.g. \cite{lin2010rank}), such as determining the ranking that minimizes the sum of the distances of the separate rankings to the \textit{consensus ranking}. As a special case, using Spearman's distance produces the consensus ranking given by 
averaging ranks (with average ranks in case of ties instead of their minimum) across tasks for each algorithm and ranking these averages. Note that each task contributes equally to the consensus ranking independent of its sample size or ranking stability unless weights are assigned to each task.
\subsection{Investigating ranking stability}\label{sec:methods:stability}
The assessment of stability of rankings across different ranking methods with respect to both sampling variability and variability across tasks (i.e. generalizability of algorithms across tasks) is of major importance~\cite{maierhein2018rankings}. This is true particularly if there is a small number of test cases.   
In this section, we will review two approaches for investigating ranking stability.
\subsubsection{Bootstrap approach}\label{sec:bootstrapapproach}
For a given ranking method, the bootstrap distribution of rankings for each algorithm may be used to assess the stability of an algorithm's ranking with respect to sampling variability. To this end, the ranking strategy is performed repeatedly on each bootstrap sample. One bootstrap sample of a task with $n$ test cases consists of $n$ test cases randomly drawn with replacement from this task. A total of $b$ of these bootstrap samples are drawn (e.g., $b=1,000$). Bootstrap approaches can be evaluated in two ways: Either the rankings for each bootstrap sample are evaluated for each algorithm, or 
the distribution of correlations or pairwise distances (see sec.~\ref{sec:methods:comparison})  between the ranking list based on the full assessment data and based on each bootstrap sample can be explored (see sec. \ref{sec:synthetic:stability}). 
 
\subsubsection{Testing approach}\label{sec:methods:stability:testing}
Another way to assess the uncertainty in rankings with respect to sampling variability is to employ pairwise significance tests that assess significant differences in metric values between algorithms. 
As this poses a multiple comparison problem leading to inflation of family-wise error rates, an adjustment for multiple testing, such as Holm's procedure, should be applied. Note that, as always, the lack of statistical significance of a difference may be due to having too few test cases and cannot be taken as evidence of absence of the difference.


\section{Visualization for single-task challenges}\label{sec:synthetic}

The visualization methods for single-task challenges can be classified into methods for visualization of the assessment data itself (sec.~\ref{sec:synthetic:raw}) and the robustness and stability of rankings (secs.~\ref{sec:synthetic:stability:method} and~\ref{sec:synthetic:stability}).
This section presents the methodology along with the relevant sample illustrations computed for the synthetic challenges described in secs.~\ref{sec:methods:example:ideal} and \ref{sec:methods:example:failureRandom}. To ensure that the presentation is clear, we have used explanatory boxes that show a basic description of each visualization method positioned directly under the corresponding sample plots. 
In all of the visualization schemes, algorithms are ordered according to a selected ranking method (here: \textit{aggregate-then-rank} using mean for aggregation). 

\subsection{Visualizing assessment data}\label{sec:synthetic:raw}
Visualization of assessment data helps us to understand the distribution of metric values for each algorithm across test cases. 

\subsubsection{Dot- and boxplots} \label{sec:dotboxplots}
The most commonly applied visualization technique in biomedical image analysis challenges are boxplots, which represent descriptive statistics for the metric values of one algorithm. These can be enhanced with horizontally jittered dots, which represent the individual metric values of each test case, as shown in Fig.~\ref{fig:synthetic:raw:boxplot}. In an ideal scenario (\textit{c\_ideal}), the assessment data is completely separated and the ranking can be inferred visually with ease. In other cases (here: 
\textit{c\_random)}, the plots are less straightforward to interpret, specifically because dot- and boxplots do not connect the values of the same test case for the different algorithms. A test case in which all of the methods perform poorly, for example, cannot be extracted visually. 
%

\subsubsection{Podium plots}\label{sec:podiumplots}
\begin{figure}
\begin{tcolorbox}[title= Podium plot]
\centering
\begin{subfigure}[b]{\figurewidth}
    \centering
    \caption{c\_ideal}\vspace{-.5em}
    \label{subfig:synthetic:raw:podium:a}
    \includegraphics[width=\figurewidth]{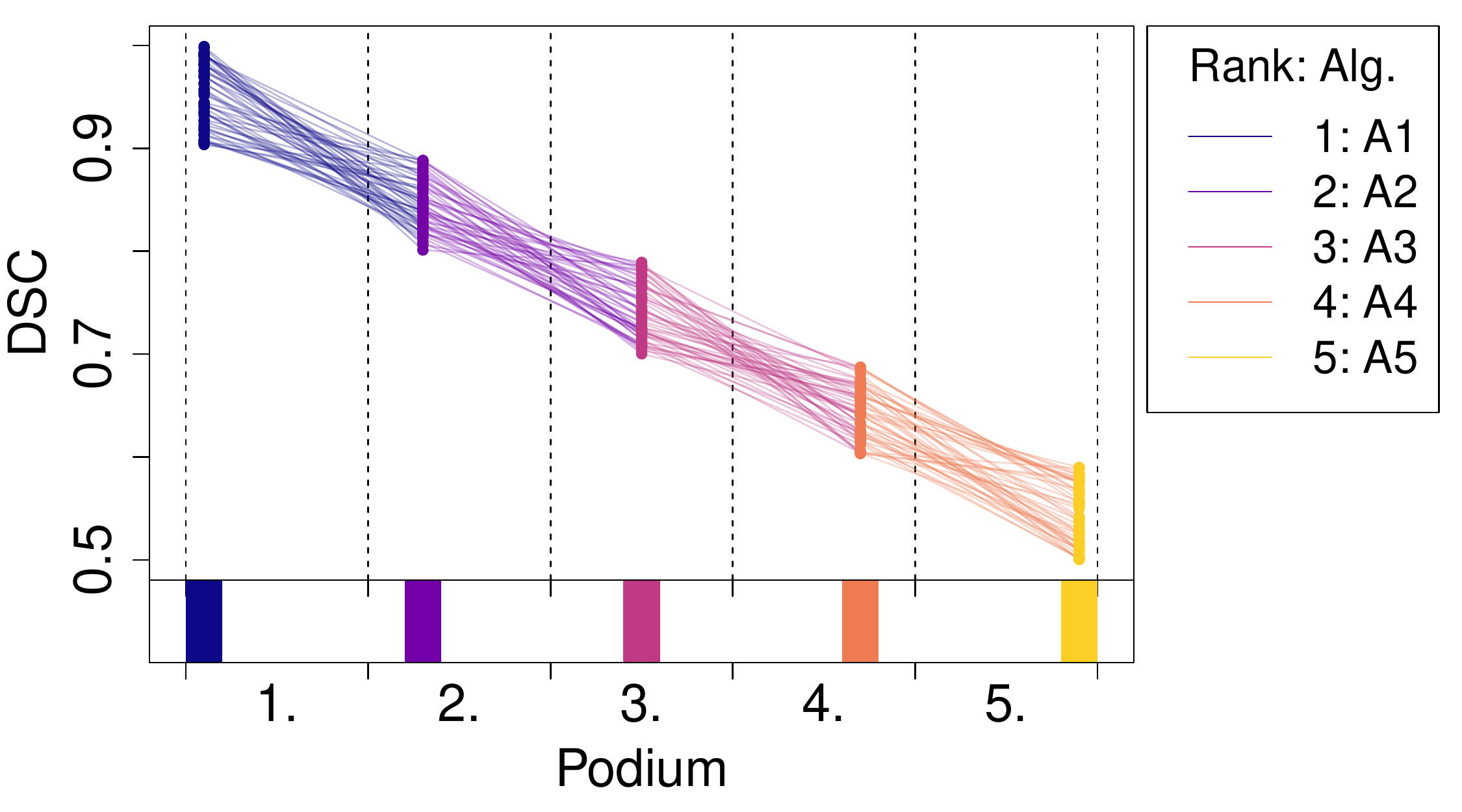}
\end{subfigure}\vspace{1em}

\begin{subfigure}[b]{\figurewidth}
    \centering
     \caption{c\_random}\vspace{-.5em}
    \label{subfig:synthetic:raw:podium:c}
   \includegraphics[width=\figurewidth]{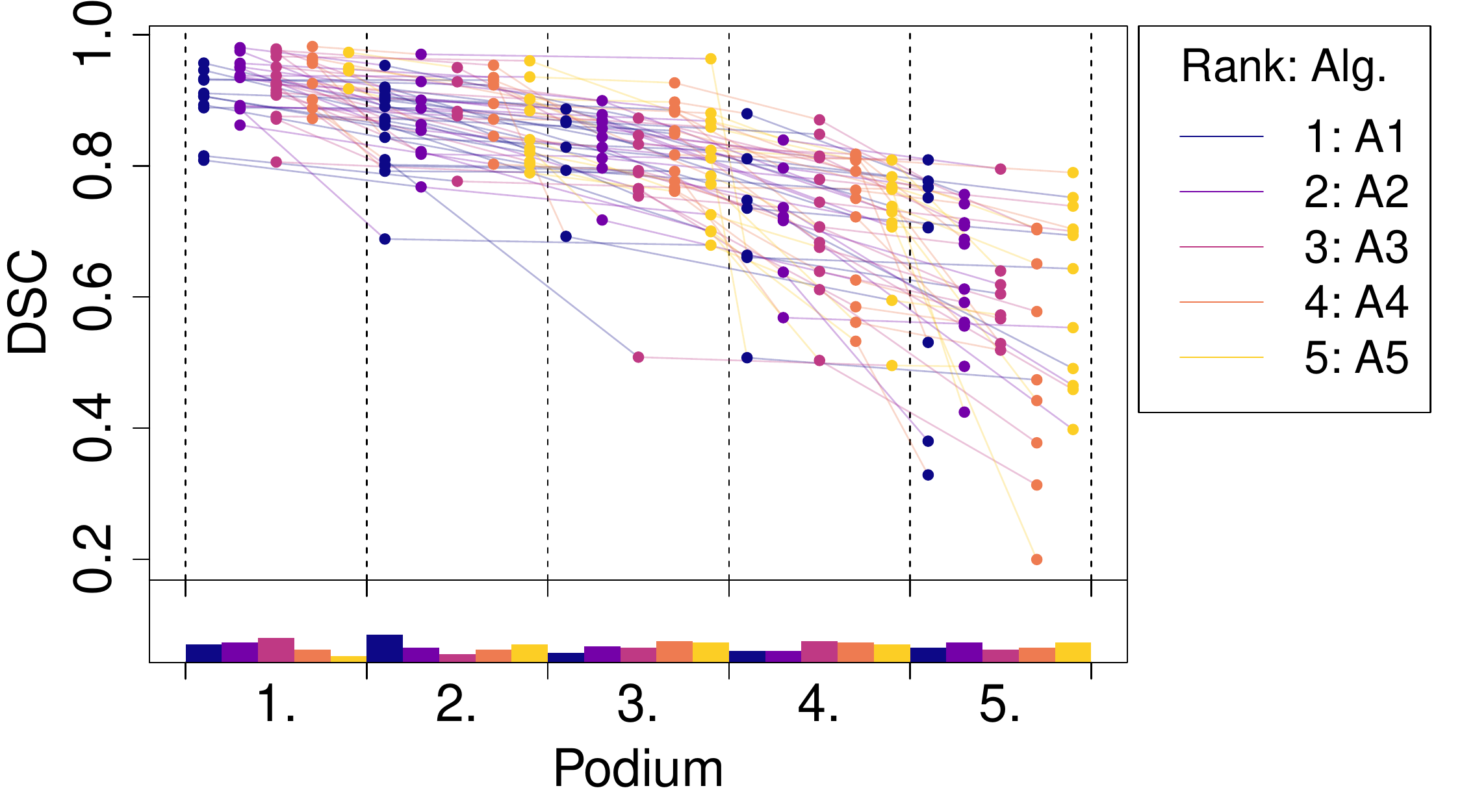} 
\end{subfigure}
\caption{\textit{Podium plots}~\cite{eugster2008exploratory} for visualizing assessment data. Upper part: Participating algorithms are color-coded, and each colored dot in the plot represents a metric value achieved with the respective algorithm. The actual metric value is encoded by the y-axis. Each podium (here: $p=5$) represents one possible rank, ordered from best (1) to worst (here: 5). The assignment of metric values (i.e. colored dots) to one of the podiums is based on the rank that the respective algorithm achieved on the corresponding test case. Note that the plot part above each podium place is further subdivided into $p$ ``columns'', where each column represents one participating algorithm.  Dots corresponding to identical test cases are connected by a line, 
producing the spaghetti structure shown here. Lower part: Bar charts represent the relative frequency at which each algorithm actually achieves the rank encoded by the podium place. 
}\label{fig:synthetic:raw:podium}
\end{tcolorbox}
\end{figure}

\textit{Benchmark experiment plots}~\cite{eugster2008exploratory}, here referred to as \textit{podium plots} overcome the well-known issues of dot- and boxplots by connecting the metric values corresponding to the same test case but different algorithms. Fig.~\ref{fig:synthetic:raw:podium} includes a description of the principle and how to read the plots. 
In an ideal challenge (\textit{c\_ideal}; 
Fig.~\ref{subfig:synthetic:raw:podium:a}), one algorithm (here:  $A_1$) has the highest metric value for all test cases. Consequently, all dots corresponding to podium place 1 share the same color (here: blue). All other ranks are represented by one algorithm and therefore one color. 
In contrast, no systematic color representation (and thus no ranking) can be visually extracted from the simulated 
random challenge, as illustrated in Fig. 
\ref{subfig:synthetic:raw:podium:c}. 
It should be mentioned that this approach requires unique ranks; in the event of ties (identical ranking for at least two algorithms), random ranks are assigned to the ties. This visualization method reaches its limit in challenges with large numbers of algorithms.

\subsubsection{Ranking heatmap}\label{sec:rankingheatmap}
\begin{figure}
\begin{tcolorbox}[title= Ranking heatmap]
\centering
\begin{subfigure}[b]{.55\figurewidth}
    \centering
    \caption{c\_ideal}\vspace{-.5em}
    \label{subfig:synthetic:raw:heatmap:a}
    \includegraphics[width=.55\figurewidth]{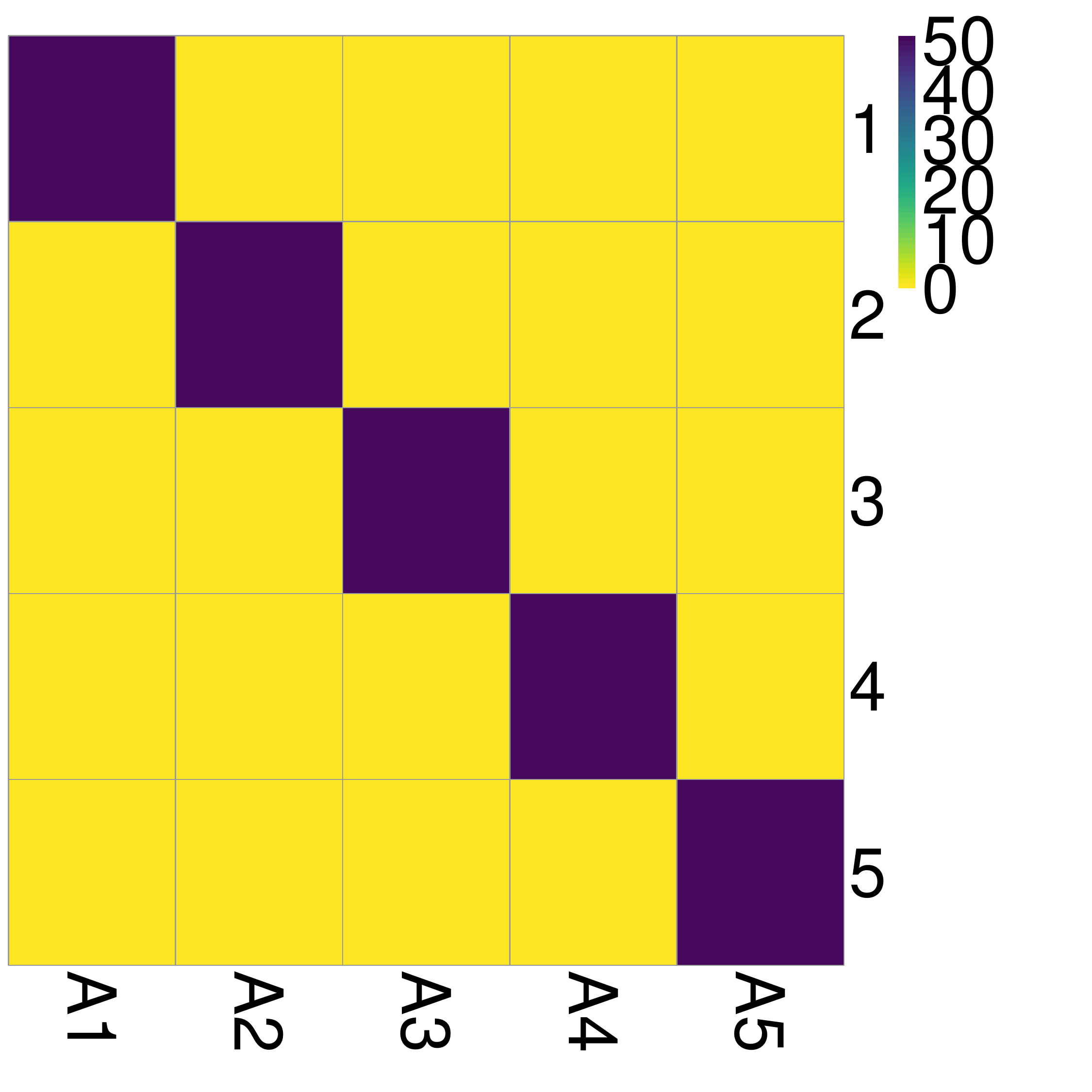}
\end{subfigure}
\begin{subfigure}[b]{.55\figurewidth}
    \centering
    \caption{c\_random}\vspace{-.5em}
    \label{subfig:synthetic:raw:heatmap:c}
    \includegraphics[width=.55\figurewidth]{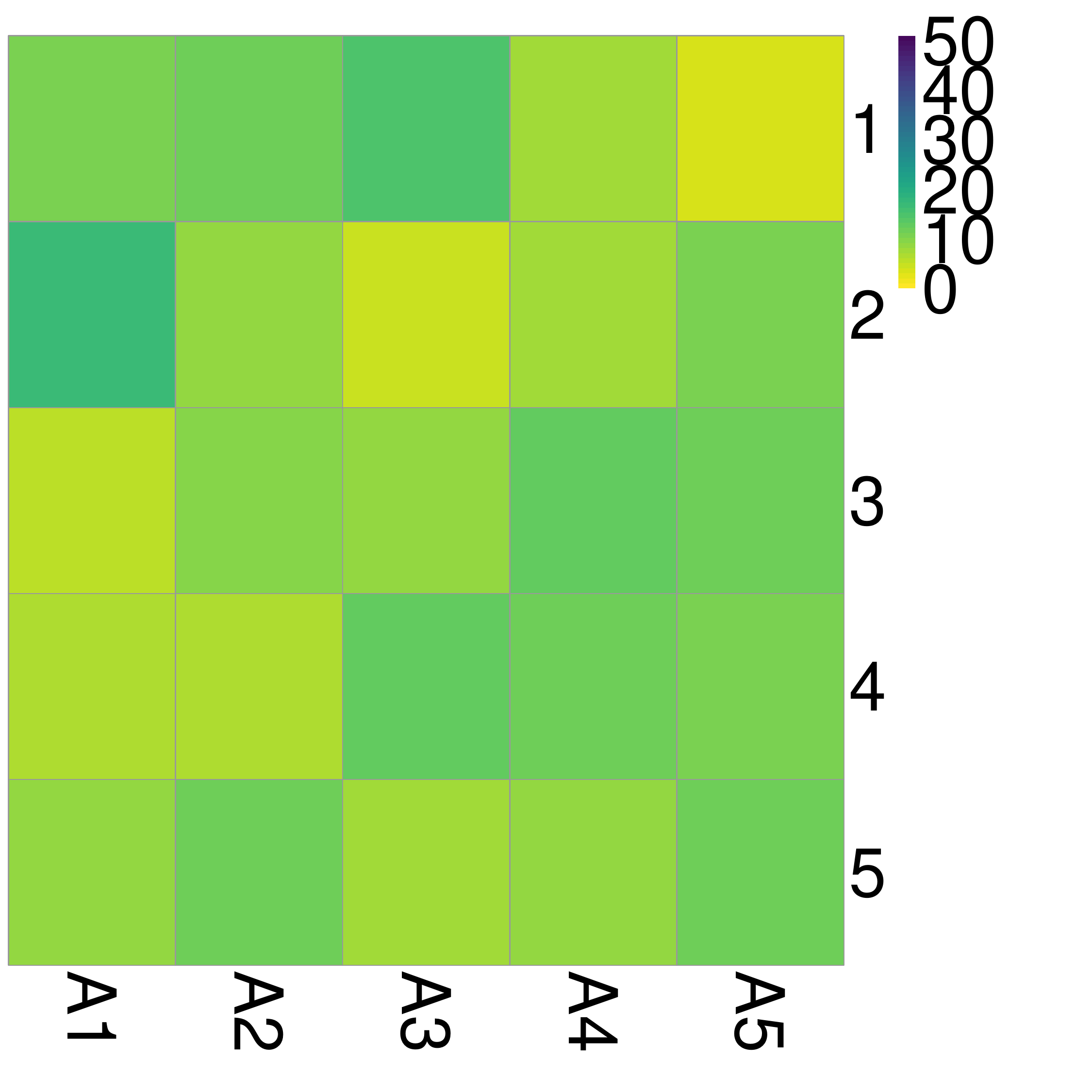} 
\end{subfigure}
\caption{\textit{Ranking heatmaps} for visualizing assessment data. Each cell $\left( i, A_j \right)$ shows the absolute frequency of test cases in which algorithm $A_j$ achieved rank $i$. }\label{fig:synthetic:raw:heatmap}
\end{tcolorbox}
\end{figure}

Another way to visualize assessment data is to use ranking heatmaps, as illustrated in Fig.~\ref{fig:synthetic:raw:heatmap}. 
These heatmaps abstract from the individual metric values and contrast rankings on a test-case basis ("rank first") to the results of the selected overall ranking method. A dark color concentrated along the diagonal indicates concordance of rankings.
In general, a higher contrast of the matrix implies better separability of algorithms. This visualization method is particularly helpful when the number of test cases is too large for an interpretable podium plot. 


\subsection{Ranking robustness with respect to ranking method} \label{sec:synthetic:stability:method}
\begin{figure}
\begin{tcolorbox}[title= Line plot for comparison of ranking methods]
\centering
\includegraphics[width=\figurewidth]{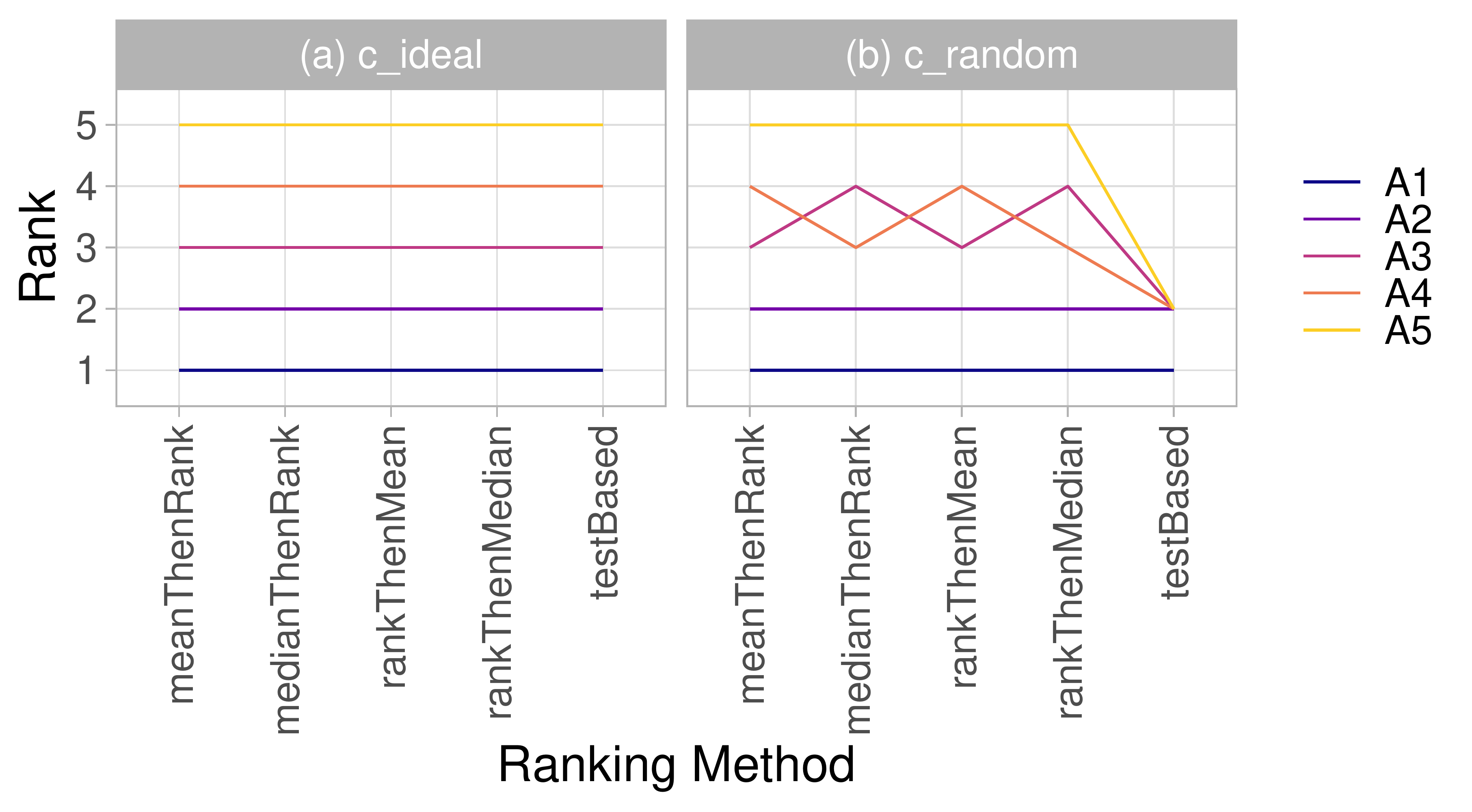} 
\caption{\textit{Line plots} for visualizing the robustness of ranking  across different ranking methods. Each algorithm is represented by one colored line. For each ranking method encoded on the x-axis, the height of the line represents the corresponding rank. Horizontal lines indicate identical ranks for all methods.}\label{fig:synthetic:stability:method}
\end{tcolorbox}
\end{figure}

Recent findings show that rankings are largely dependent on the ranking method applied~\cite{maierhein2018rankings}. One could argue, however, that if a challenge separates algorithms well, then any ranking method reflecting the challenge goal should yield the same ranking. We propose using line plots, presented in Fig.~\ref{fig:synthetic:stability:method}, to investigate this aspect for a given challenge. In an ideal scenario (Fig.~\ref{fig:synthetic:stability:method}, left), all of the lines are parallel.
In other instances, crossing lines indicate sensitivity to the choice of the ranking method.

\subsection{Ranking stability for a selected ranking method}\label{sec:synthetic:stability}
In sec.~\ref{sec:methods:stability}, we identified two basic means for investigating ranking stability: bootstrapping and the testing approach. This section describes different ways to present the data resulting from these analyses. 

\subsubsection{Visualizing bootstrap results}  \label{sec:synthetic:stability:bootstrap}

\begin{figure}
\begin{tcolorbox}[title= Blob plot]
\centering
    \centering
    \includegraphics[width=\figurewidth]{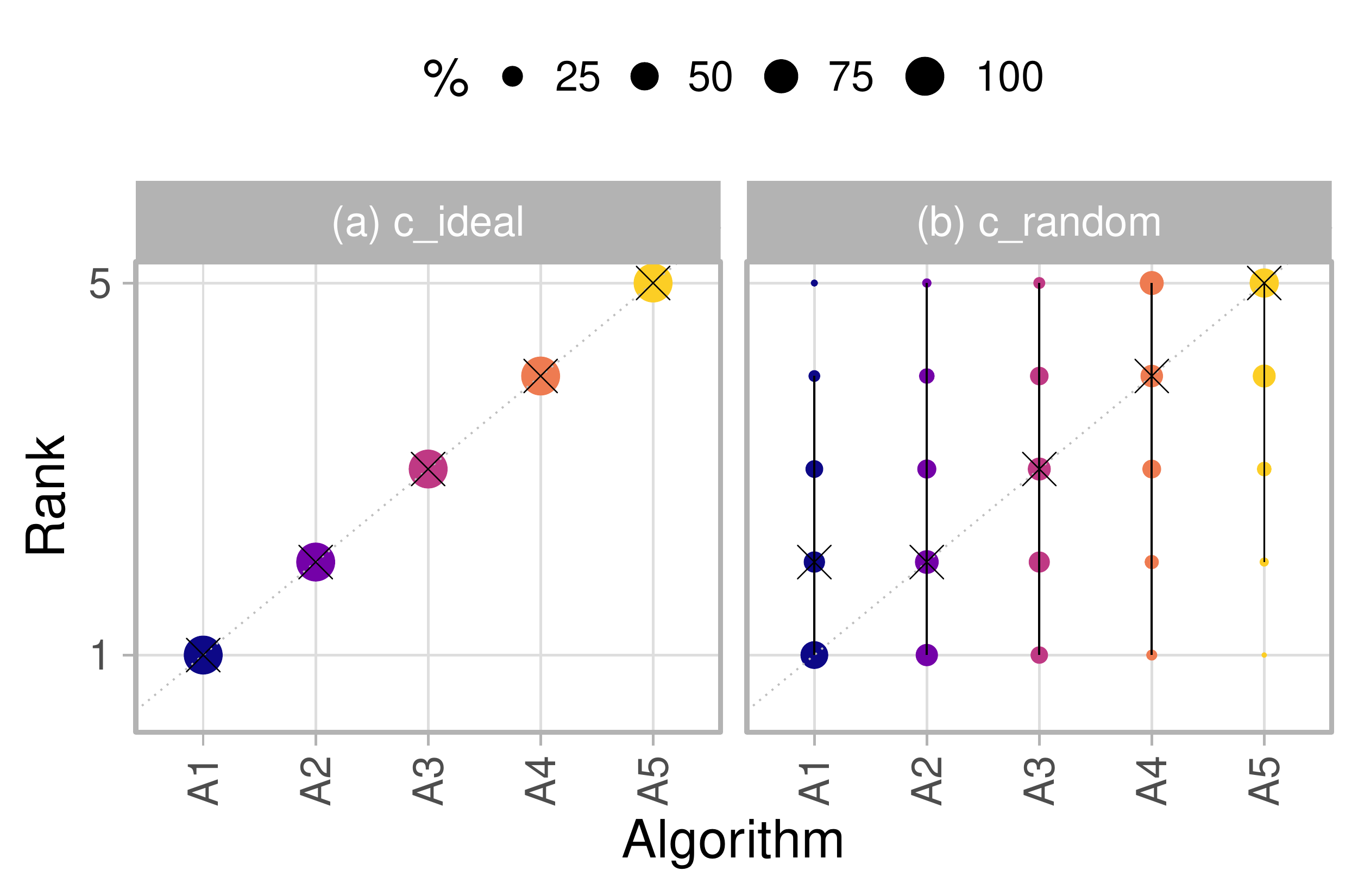}
\caption{\textit{Blob plots} for visualizing ranking stability 
based on
bootstrap 
sampling.
Algorithms are color coded, and the area of each blob at position $\left( A_i, \text{rank } j \right)$ is proportional to the relative frequency $A_i$ achieved rank $j$ (here across $b=1000$ bootstrap samples). The median rank for each algorithm is indicated by a black cross. 
95\% bootstrap intervals across bootstrap samples (ranging from the 2.5th to the 97.5th percentile of the bootstrap distribution) are indicated by black lines. }\label{fig:synthetic:stability:bootstrap}
\end{tcolorbox}
\end{figure}

An intuitive way to comprehensively visualize bootstrap results are \textit{blob plots}, as illustrated in Fig.~\ref{fig:synthetic:stability:bootstrap}.  
As the existence of a blob requires an absolute frequency of at least one, a small number of blobs typically indicates higher certainty, as illustrated in Fig.~\ref{fig:synthetic:stability:bootstrap}a. In contrast, many blobs of comparable size suggest high uncertainty, see Fig.~\ref{fig:synthetic:stability:bootstrap}b. 

\begin{figure}
\begin{tcolorbox}[title= Violin plot]
\centering
\includegraphics[width=\figurewidth]{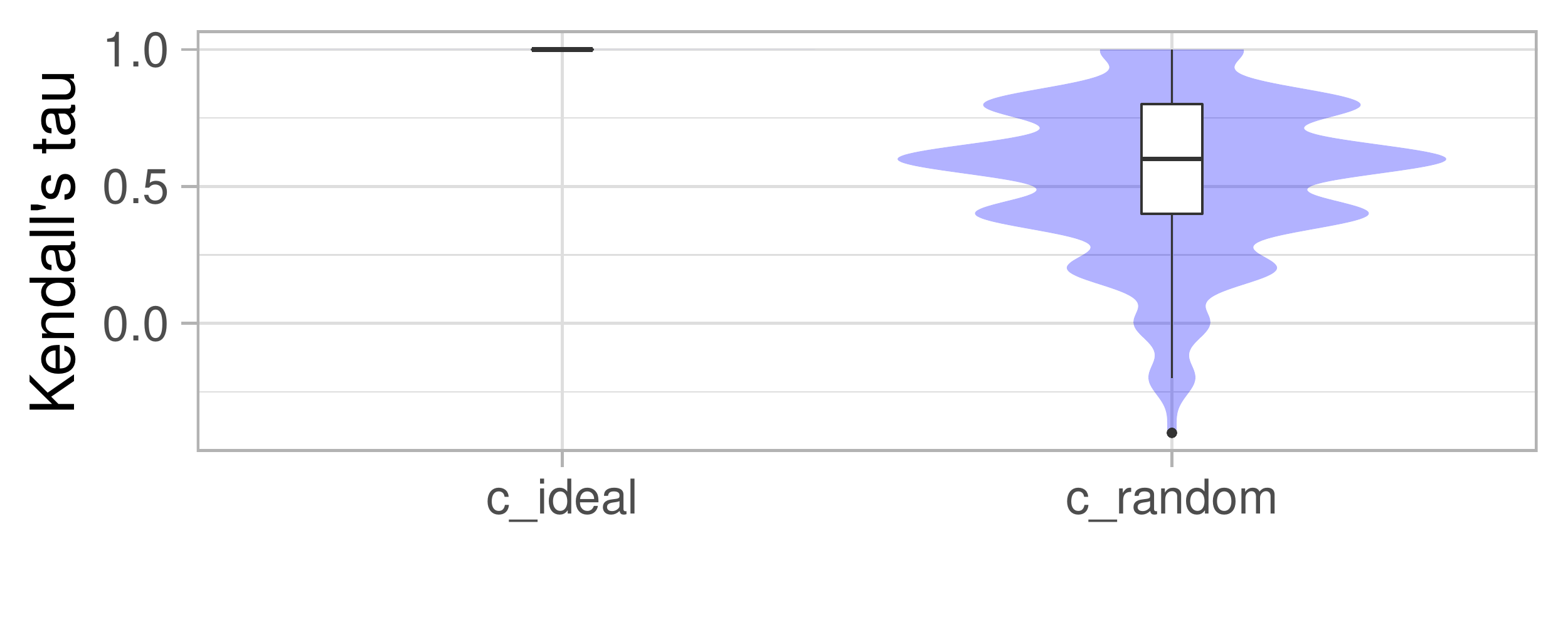} 
\caption{\textit{Violin plots} for visualizing ranking stability based on bootstrapping. The ranking list based on the full assessment data is compared pairwise with the ranking lists based on the individual bootstrap samples (here $b=1000$ samples). Kendall's $\tau$ (cf. sec.~\ref{sec:methods:comparison}) is computed for each pair of rankings, and a violin plot that simultaneously depicts a boxplot and a density plot is generated from the results.}\label{fig:synthetic:stability:violin}
\end{tcolorbox}
\end{figure}

\textit{Violin plots}, as shown and described in Fig.~\ref{fig:synthetic:stability:violin}, provide a 
more condensed 
way to analyze bootstrap results. In these plots, the focus is on the comparison of the ranking list 
computed
on the full assessment data and 
the individual bootstrap samples, respectively. Kendall's $\tau$ is chosen for comparison as it is has an upper and lower bound (+1/-1). In an ideal scenario (here \textit{c\_ideal} ), the ranking is identical to the full assessment data ranking in each bootstrap sample. Hence, Kendall's $\tau$ is always equal to one, demonstrating perfect stability of the ranking. In \textit{c\_random}, values of Kendall's $\tau$ are very dispersed across the bootstrap samples, indicating high instability of the ranking. 

\subsubsection{Testing approach summarized by significance map}\label{sec:synthetic:stability:significance}
\begin{figure}
\begin{tcolorbox}[title= Significance map]
\centering
\begin{subfigure}[b]{.5\figurewidth}
    \centering
    \caption{c\_ideal}\vspace{-.5em}
   \includegraphics[width=.55\figurewidth]{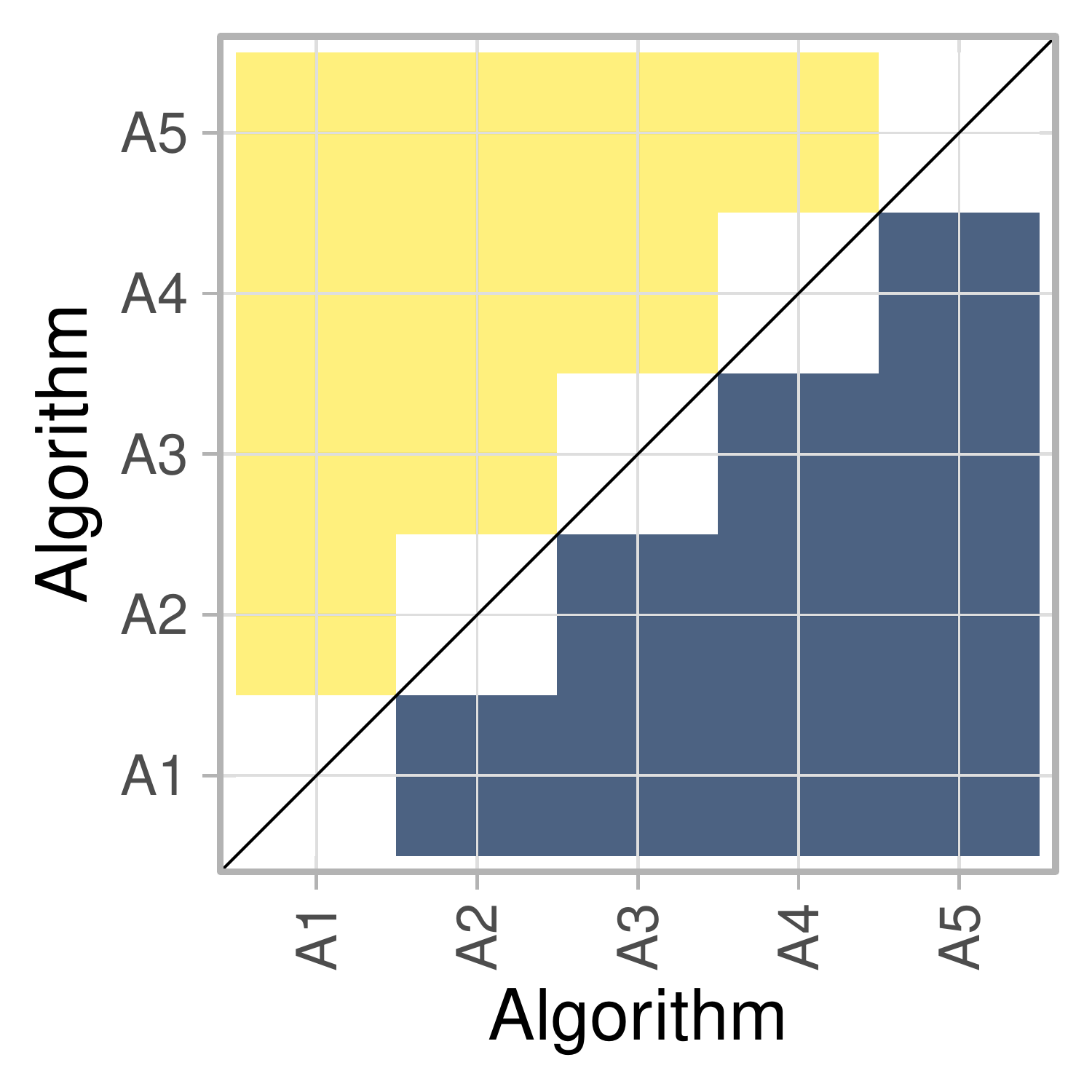}
     \label{subfig:synthetic:stability:significance:a}
\end{subfigure}
\begin{subfigure}[b]{.5\figurewidth}
    \centering
   \caption{
   c\_random}\vspace{-.5em}
    \includegraphics[width=.55\figurewidth]{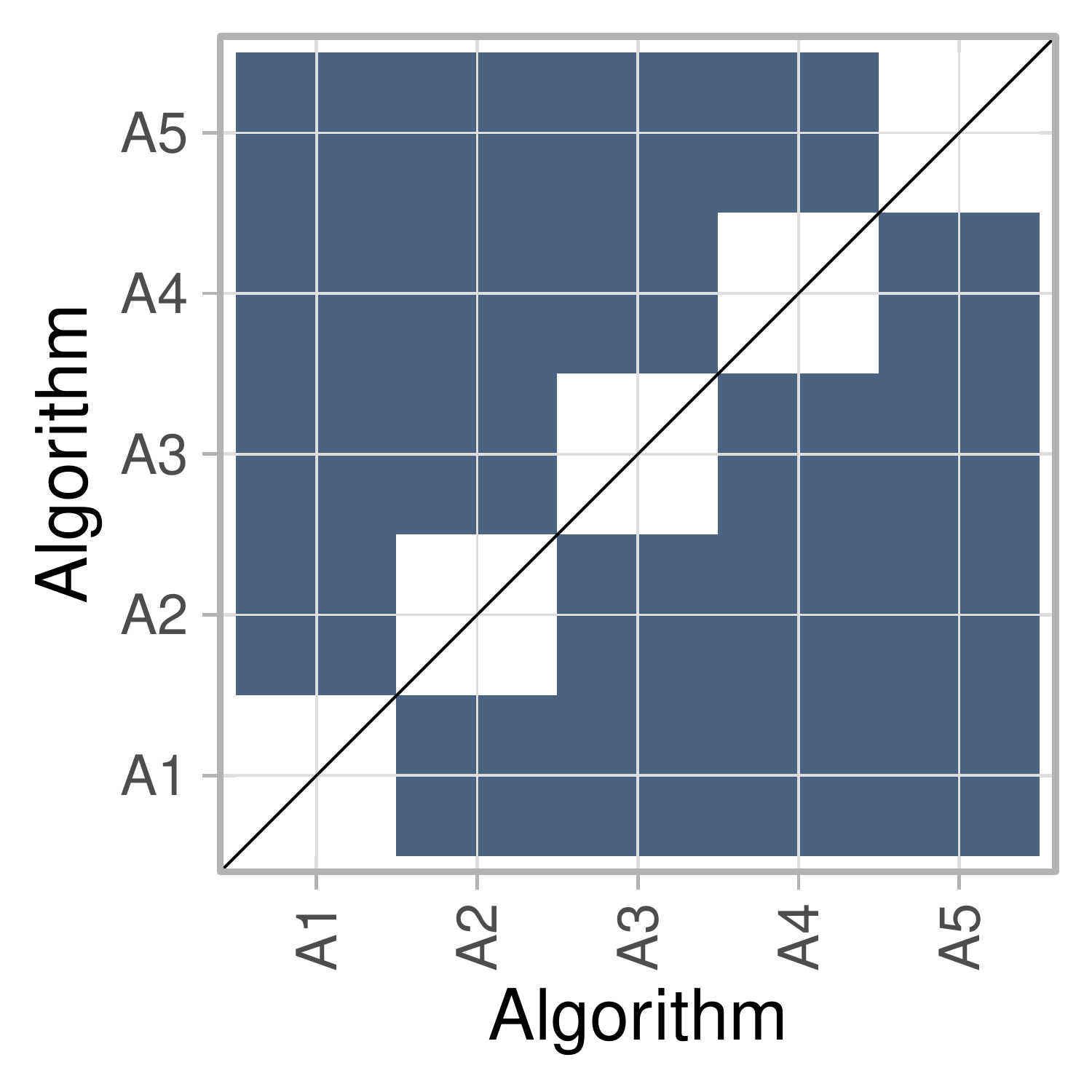}
     \label{subfig:synthetic:stability:significance:b}
\end{subfigure}
\caption{\textit{Significance maps} for visualizing ranking stability based on statistical significance. 
They depict incidence matrices of
pairwise significant test results for the one-sided Wilcoxon signed rank test at 5\% significance level with adjustment for multiple testing according to Holm. 
Yellow
shading indicates that metric values of the algorithm on the x-axis are significantly superior to 
those from the algorithm on the y-axis, 
blue
color indicates no significant difference.}\label{fig:synthetic:stability:significance}
\end{tcolorbox}
\end{figure}

As described in sec.~\ref{sec:methods:stability:testing}, an alternative way to assess ranking stability is significance testing. To visualize the pairwise significant superiority between algorithms, we propose the generation of a \textit{significance map}, as illustrated in~Fig.~\ref{fig:synthetic:stability:significance}. In an ideal scenario (\textit{c\_ideal}), ordering is optimal and all algorithms with smaller rank are significantly better than algorithms with larger rank, leading to a yellow 
area above and a blue 
area below the diagonal, respectively. The high uncertainty in 
\textit{c\_random}
is reflected by the uniform blue 
color. 


\section{Visualization for multi-task challenges}\label{sec:synthetic:multiple}
Several challenges comprise multiple tasks. A common reason for this is that a clinical problem may involve solving several sub-problems, each of which is relevant to the overall goal. Furthermore, single-task challenges do not allow us to investigate how algorithms generalize to different tasks. 
This section is devoted to the visualization of the important characteristics of algorithms (sec.~\ref{sec:synthetic:multiple:stability}) and tasks (sec.~\ref{sec:synthetic:multiple:task}) in such multi-task challenges. As most methods are based on the concepts presented in the previous section, the illustration is performed directly with real world data (see sec.~\ref{sec:methods:example:real}). Algorithms are ordered according to a consensus ranking (see sec.~\ref{sec:methods:comparison}) based on average ranks across tasks.

\subsection{Characterization of algorithms} \label{sec:synthetic:multiple:stability}
In most multi-task challenges, algorithms are in the focus of the analysis. The primary goal is to identify methods that consistently outperform competing algorithms across all tasks. We propose two methods for analyzing this:

\subsubsection{Visualization of ranking variability across tasks}\label{sec:synthetic:multiple:stability:rankingacrosstasks}
If a reasonably large number of tasks is available, a blob plot similar to the one shown in Fig.~\ref{fig:synthetic:stability:bootstrap} can be drawn by substituting rankings based on bootstrap samples with the rankings based on multiple tasks. This way, the distribution of ranks across tasks can be intuitively visualized as shown in Fig.~\ref{fig:realMultiple:stability:raw}. 
All ranks that an algorithm achieved in any task are displayed along the y-axis, with the area of the blob being proportional to the frequency. 
If all tasks provided the same stable ranking, narrow intervals around the diagonal would be expected.

\subsubsection{Visualization of ranking variability based on bootstrapping}\label{sec:synthetic:multiple:stability:bootsrtap} \label{sec:synthetic:multiple:stability:bootstrap}
A variant of the blob plot approach illustrated in Fig.~\ref{fig:synthetic:stability:bootstrap} involves replacing the algorithms on the x-axis with the tasks and then generating a separate plot for each algorithm as shown in  Fig.~\ref{subfig:realMultiple:stability:bootstrap:a}. This allows assessing the variability of rankings for each algorithm across multiple tasks and bootstrap samples.
Here, color coding is used for the tasks, and separation by algorithm enables a relatively straightforward strength-weaknesses analysis for individual methods. 



%
%

\subsection{Characterization of tasks} \label{sec:synthetic:multiple:task}

It may also be useful to structure the analysis around the different tasks. This section proposes visualization schemes to analyze and compare tasks of a competition.
\subsubsection{Visualizing bootstrap results}\label{sec:synthetic:multiple:task:bootstrap}

Two visualization methods are recommended to investigate which tasks separate algorithms well (i.e. lead to a stable ranking). 
Bootstrap results can be shown per task in a blob plot similar to the one described in sec.~\ref{sec:synthetic:stability:bootstrap}. Algorithms should be ordered according to the consensus ranking (Fig.~\ref{subfig:realMultiple:stability:bootstrap:b}). 
In this graph, 
tasks leading to stable (unstable) rankings are indicated by narrow (wide) spread of the blobs for all algorithms.
%

Again, to obtain a more condensed visualization,
violin plots (as presented in Fig.~\ref{fig:synthetic:stability:violin}) can be applied separately to all tasks (Fig.~\ref{fig:realMultiple:task:bootstrap:violin}). 
The overall stability of the rankings can then be compared by assessing the locations and lengths of the violins.

\subsubsection{Cluster analysis}\label{sec:synthetic:multiple:task:cluster}
\begin{figure}
\begin{tcolorbox}[title= Hierarchical clustering and networks]
\centering
\begin{subfigure}[b]{\figurewidth}
    \centering
    \caption{Dendrogram~~~~~~~~(b) Network-type graph}\vspace{-.5em}
    \label{subfig:realMultiple:task:clustering:a}
    \includegraphics[width=.49\figurewidth]{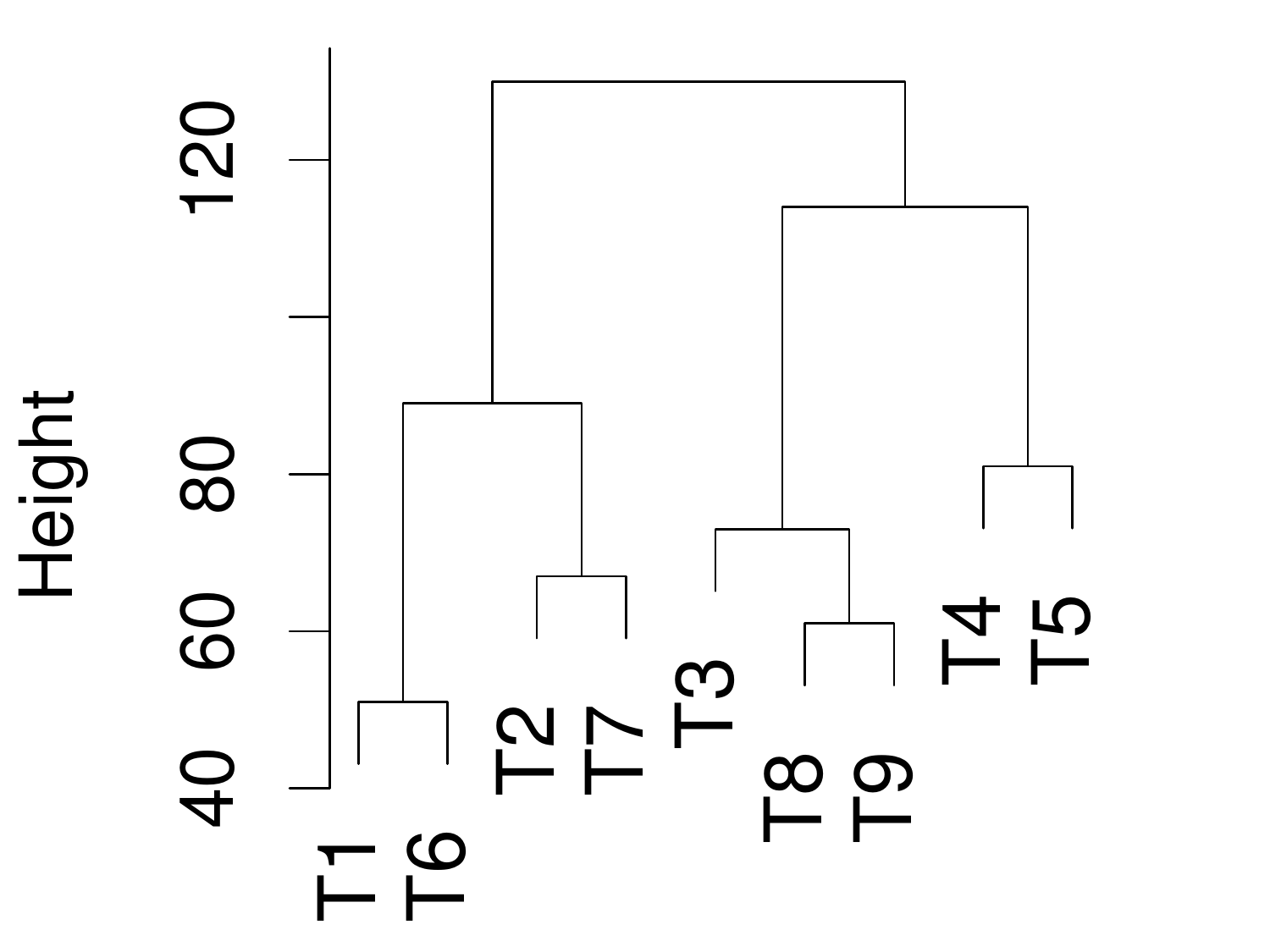} 
%
\includegraphics[width=0.49\figurewidth]{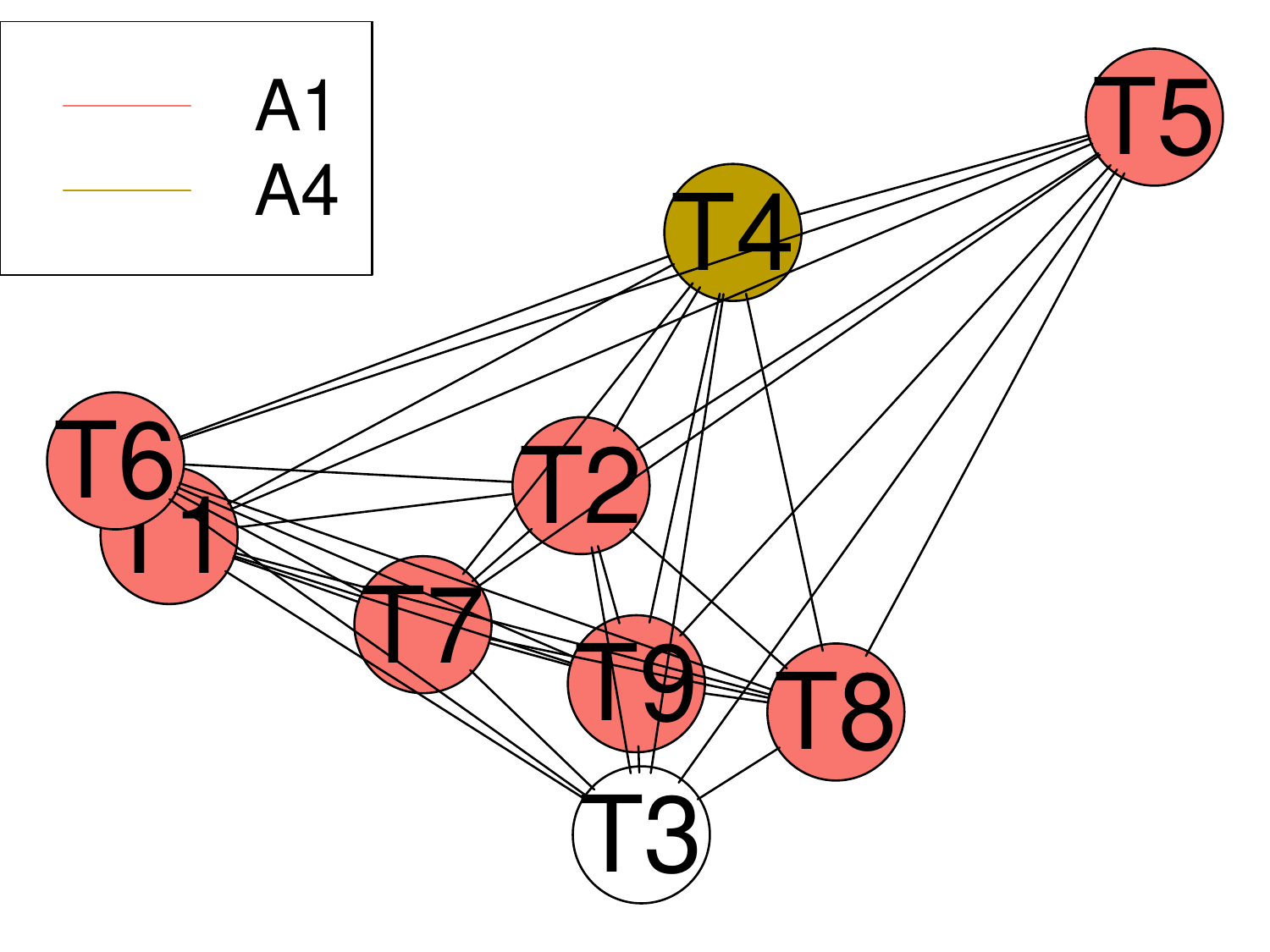} 
\end{subfigure}
\caption{\textit{Dendrogram from hierarchical cluster analysis} (a) and \textit{network-type graphs} (b) for assessing the similarity of tasks based on challenge rankings. A dendrogram (a) is a visualization approach based on hierarchical clustering, a method comprehensively described in \cite{hastie2009}. It depicts clusters according to a distance measure (here: Spearman's footrule (see sec.~\ref{sec:methods:comparison})) and an agglomeration method (here: complete agglomeration). In network-type graphs (b) \cite{eugster2008exploratory}, every task is represented by a node, and nodes are connected by edges, the length of which is determined by a distance measure (here: Spearman's footrule). Hence, tasks that are similar with respect to their algorithm ranking appear closer together than those that are dissimilar. Nodes representing tasks with a unique winner are color coded by the winning algorithm. If there is more than one first-ranked algorithm in a task, the corresponding node remains uncolored.
}\label{fig:realMultiple:task:clustering}
\end{tcolorbox}
\end{figure}

There is increasing interest in assessing the similarity of the tasks, e.g., for pre-training a machine learning algorithm. A potential approach to this could involve the comparison of the rankings for a challenge. Given the same teams participate in all tasks, 
it may be of interest to cluster tasks into groups where rankings of algorithms are similar and to identify tasks which lead to very dissimilar rankings of algorithms.
To enable such an analysis, we propose the generation of a \textit{dendrogram from hierarchical cluster analysis} or a \textit{network-type graph}, see Fig.~\ref{fig:realMultiple:task:clustering}.

\section{Open-source challenge visualization toolkit}
\label{sec:toolkit}
\begin{figure}
\centering
\includegraphics[width=\figurewidth]{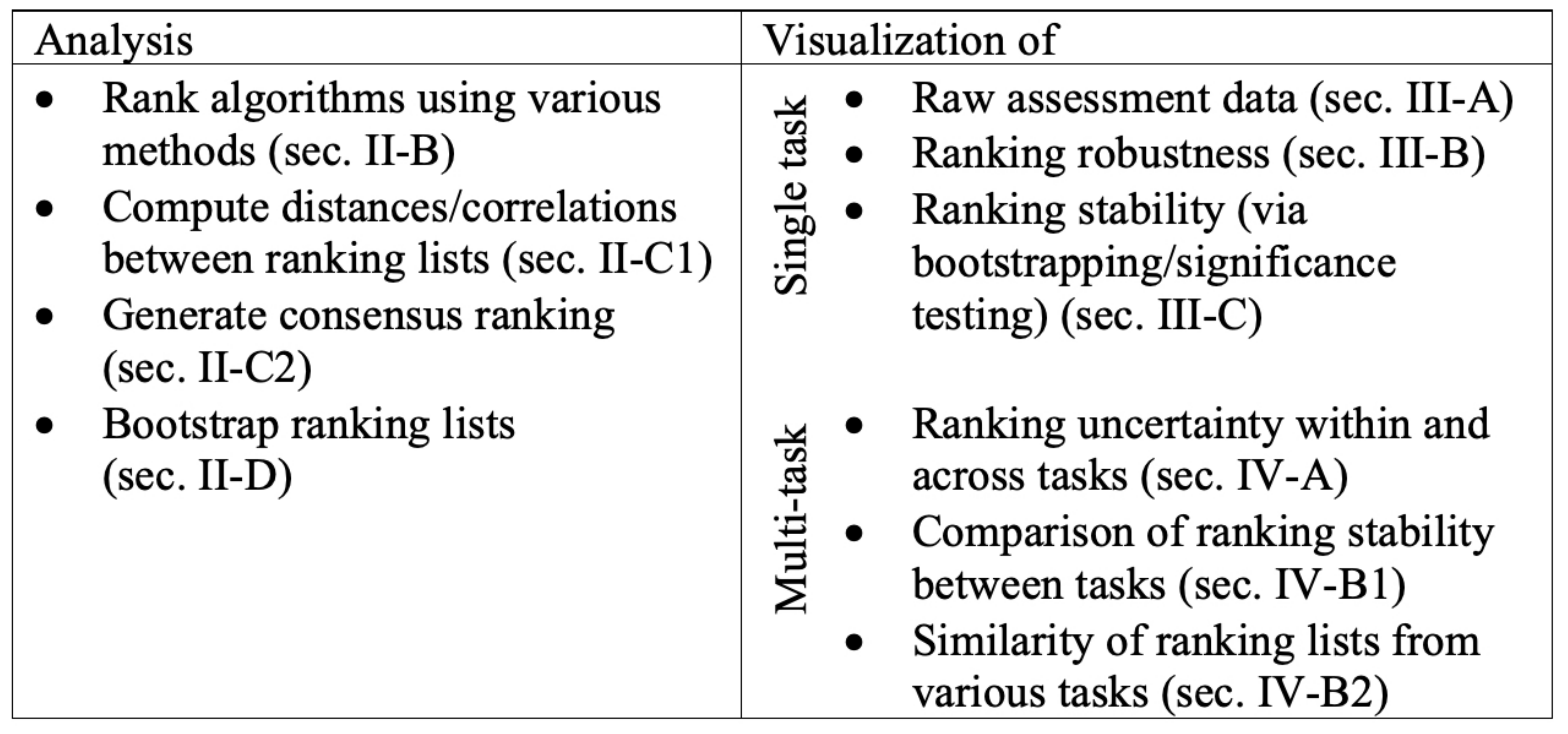} 
\caption{\textit{challengeR} as a toolkit for challenge analysis and visualization: Summary of functionality.}\label{fig:framework}
\end{figure}

All analysis and visualization methods presented in this work have been implemented in R and are provided to the community as open-source framework \textit{challengeR}. Fig. \ref{fig:framework} summarizes the functionality of the framework. The framework also offers a tool for generating full analysis reports, when it is provided with the assessment data of a challenge (csv file with columns for the metric values,
the algorithm names, test case identifiers and task identifiers in case of multi-task challenges).
Details on the framework can be found on \url{https://github.com/wiesenfa/challengeR}.

\section{Results}

To assess the applicability of our toolkit, we applied it to a recently conducted multi-task challenge (cf. sec.~\ref{sec:methods:example:real}) involving 19 participating algorithms and 17 different (sub-) tasks. Due to length restrictions, we limited the illustration of single-task visualization tools to two selected tasks:  $T_1$, which has many test cases and a relatively clear ranking, and task $T_2$, which has a small number of test cases and a more ambiguous ranking. 
1000 bootstrap samples were drawn to assess ranking variability.

\subsection{Visualization of results per task}\label{sec:realSingle}
In all of the plots, the algorithms are ordered by a test-based procedure (called \textit{significance ranking} in the following) for the specific task, performed based on the one-sided Wilcoxon signed rank test at 5\% significance level. 

\subsubsection{Visualization of assessment data}\label{sec:realSingle:raw}
\begin{figure}
\centering
\begin{subfigure}[b]{\figurewidth}
    \centering
    \caption{Task $T_1$}\vspace{-.5em}
    \label{subfig:realSingle:boxplot:a}
   \includegraphics[width=.95\figurewidth]{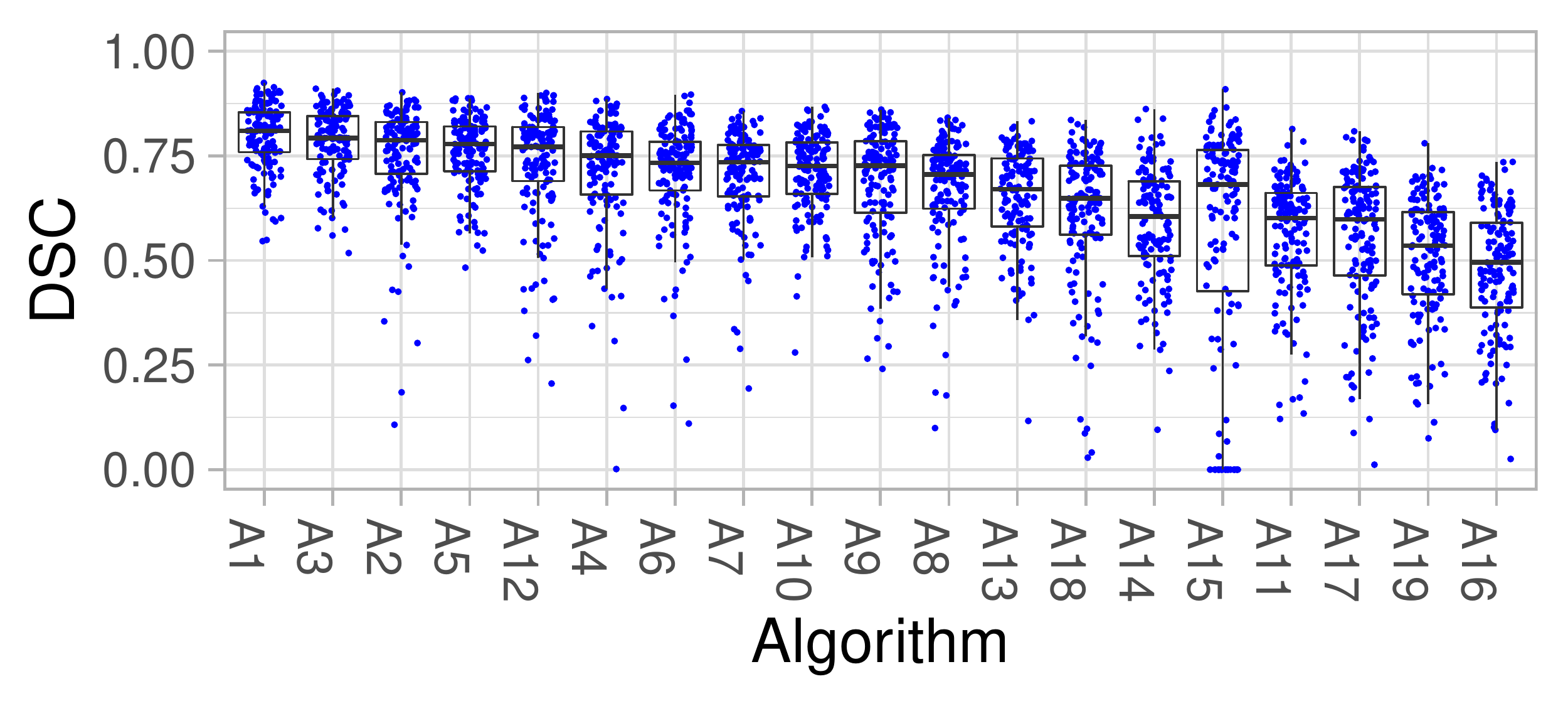}
 \end{subfigure}
 
 \vspace{1em}
\begin{subfigure}[b]{\figurewidth}
    \centering
     \caption{Task $T_2$}\vspace{-.5em}
    \label{subfig:realSingle:boxplot:b}
   \includegraphics[width=.95\figurewidth]{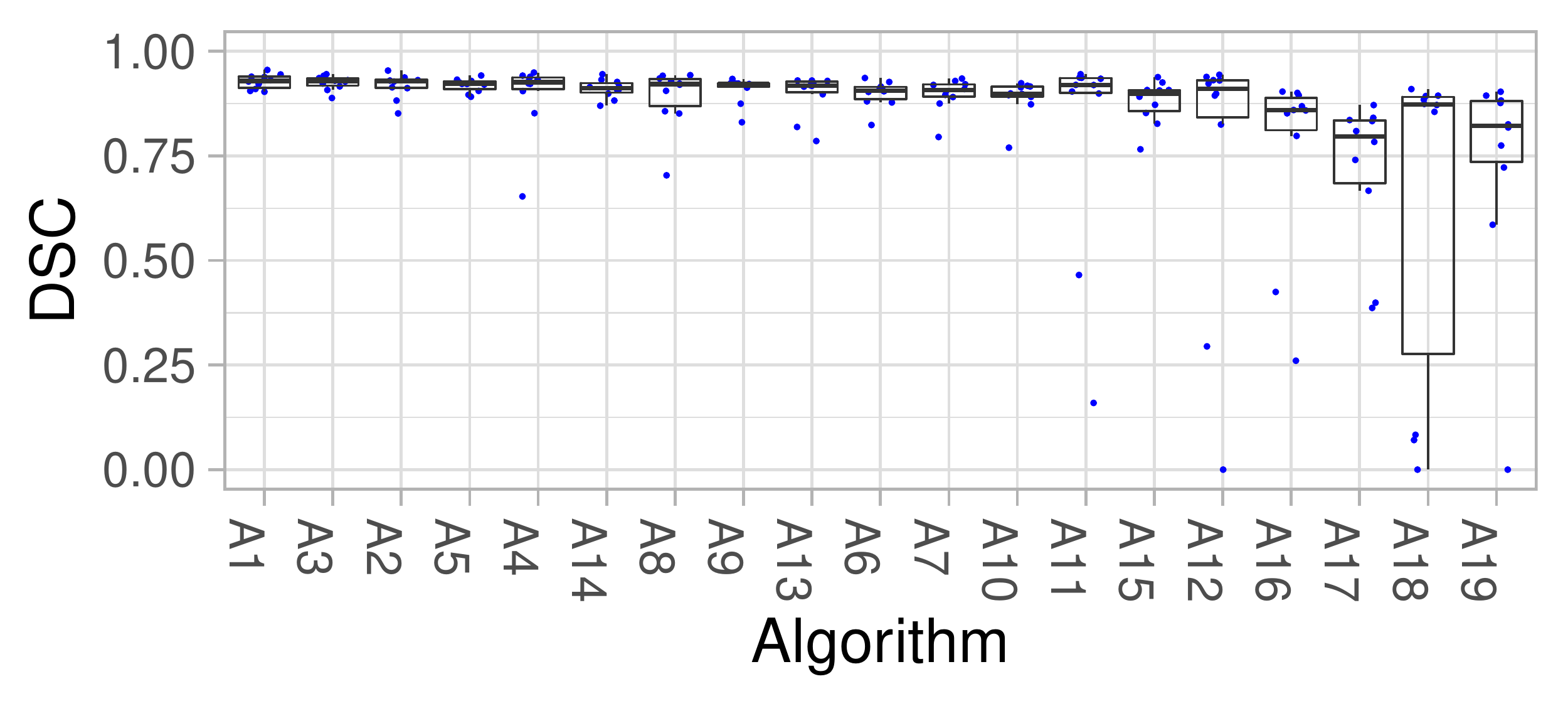}
\end{subfigure}
\caption{Dot- and boxplots visualize the raw assessment data for selected tasks of the MSD. 
}\label{fig:realSingle:boxplot}
\end{figure}

The dot- and boxplots for task $T_1$ (Fig.~\ref{subfig:realSingle:boxplot:a}) show a large number of test cases, and the quartiles suggest a relatively clear ordering. This is far less evident in Fig.~\ref{subfig:realSingle:boxplot:b} for task $T_2$, which only contains ten test cases and almost perfect metric values of most algorithms. In both tasks, a number of outliers are obvious but it remains unclear whether they correspond to the same test cases. 

\begin{figure}
\centering
\begin{subfigure}[b]{\figurewidth}
    \centering
     \caption{Task $T_1$}\vspace{-.5em}
    \label{subfig:realSingle:podium:a}
   \includegraphics[width=.98\figurewidth]{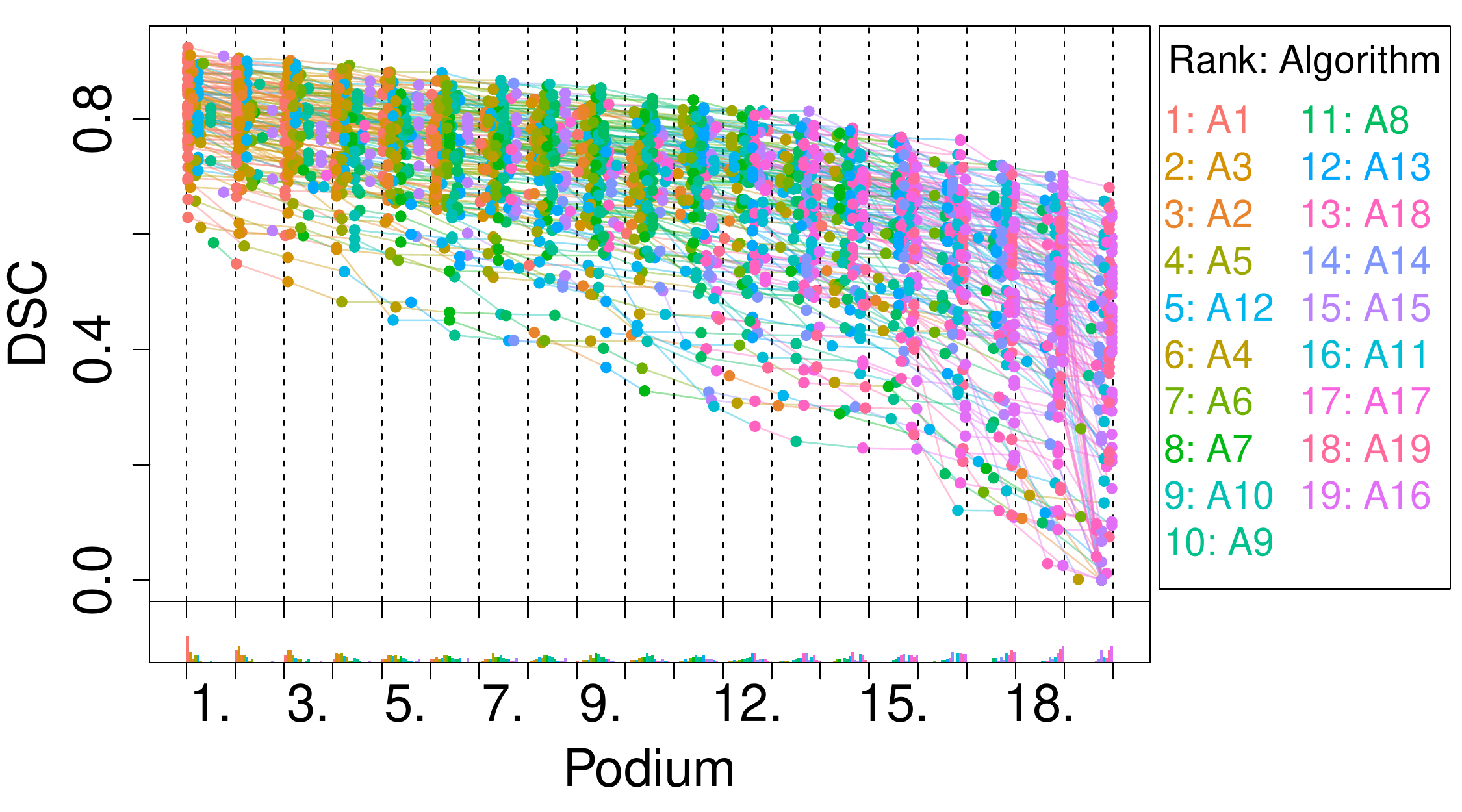}
\end{subfigure}

\vspace{1em}
\begin{subfigure}[b]{\figurewidth}
    \centering
      \caption{Task $T_2$}\vspace{-.5em}
    \label{subfig:realSingle:podium:b}
  \includegraphics[width=.98\figurewidth]{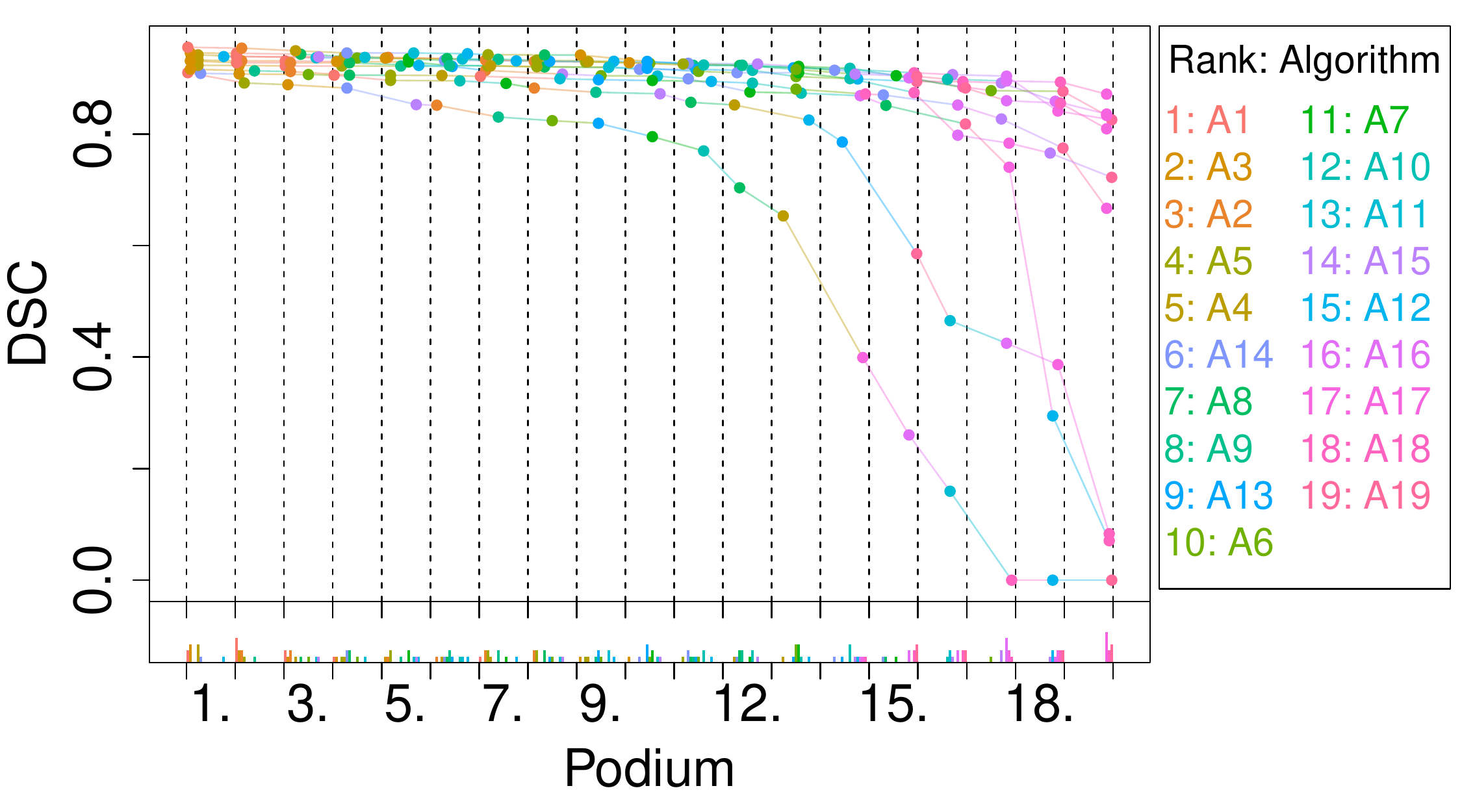}
\end{subfigure}
\caption{Podium plots visualize the assessment data for selected tasks of the MSD. $T_1$/$T_2$: task with stable/unstable ranking.   }\label{fig:realSingle:podium}
\end{figure}

In the podium plot for $T_1$ (Fig.~\ref{fig:realSingle:podium}), both the color pattern of the lines and the bar charts suggest a clear ranking for the best and the worst algorithms.
The first ranked algorithm, $A_1$, was among the first three best performing algorithms for almost all test cases. The fifth-last ranked algorithm ($A_{15}$) did not submit a valid segmentation result in numerous test cases, and hence these DSC values were set to 0, resulting in a high frequency at podium place 19. All other algorithms provided a valid value, which could be deduced from the often steep decline of the lines that end in the point corresponding to $A_{15}$ with DSC$=0$. 
%
The podium plot for $T_2$ (Fig.~\ref{subfig:realSingle:podium:b}) shows that many of the algorithms perform similarly for most of the test cases. Evidently, the assessment data were not sufficient to determine a clear ranking of the algorithms. Intriguingly, there are three test cases where algorithms perform very differently, and final rankings might be strongly affected by these test cases given the small number of test cases for this task.

\begin{figure}
\centering
\begin{subfigure}[b]{.49\figurewidth}
    \centering
    \caption{Task $T_1$}\vspace{-.5em}
    \label{subfig:realSingle:rankingHeatmap:a}
    \includegraphics[width=.55\figurewidth]{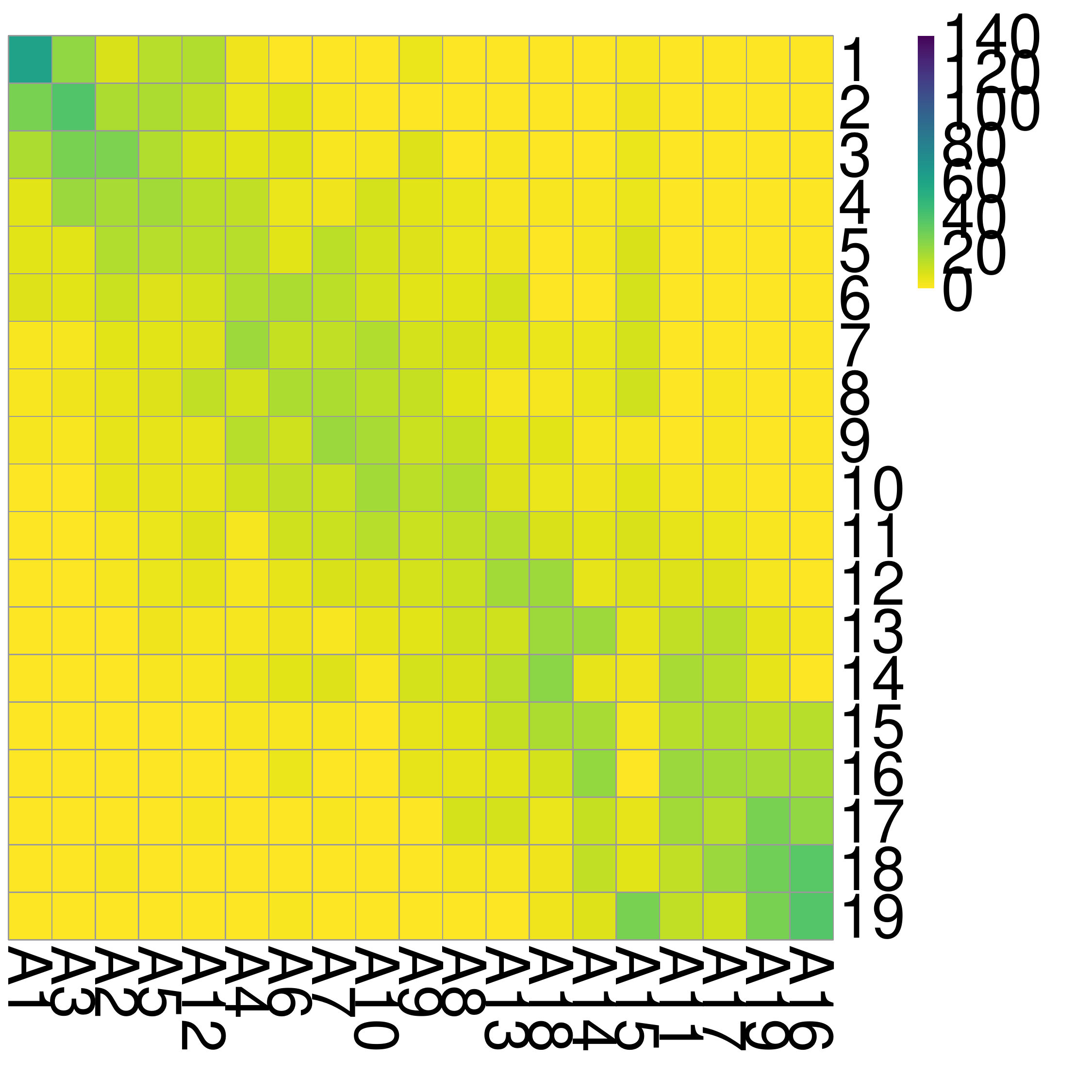}
\end{subfigure}~~~
\begin{subfigure}[b]{.49\figurewidth}
    \centering
    \caption{Task $T_2$}\vspace{-.5em}
    \label{subfig:realSingle:rankingHeatmap:b}
    \includegraphics[width=.55\figurewidth]{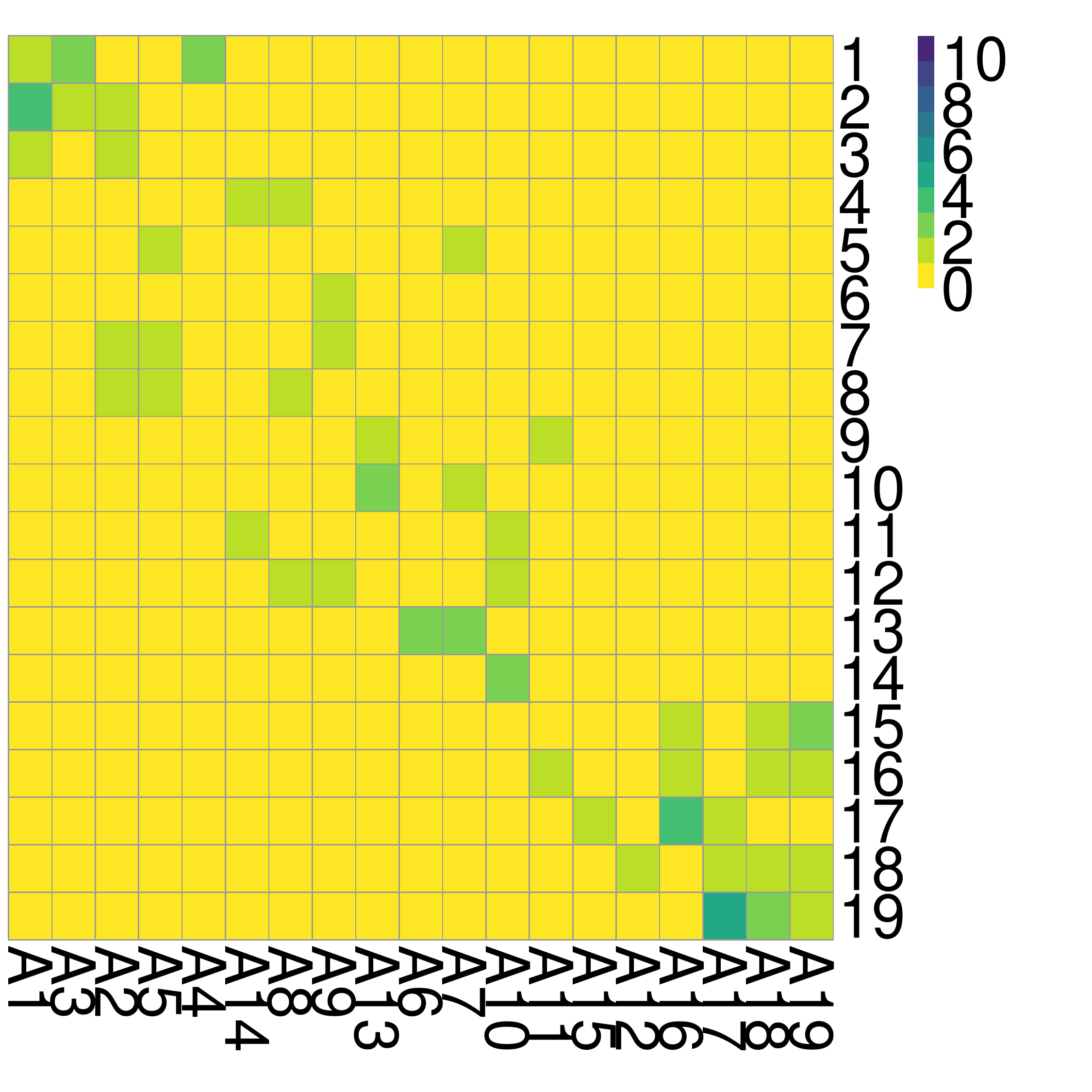}
\end{subfigure}
\caption{Ranking heatmaps for selected tasks of the MSD display the assessment data.
}\label{fig:realSingle:rankingHeatmap}
\end{figure}

Finally, Fig.~\ref{fig:realSingle:rankingHeatmap} shows the assessment data in the ranking heatmap. A relatively clear 
diagonal is observed in the left panel for task $T_1$, and this underlines the stable ranking. The right panel shows a more diverse picture with test cases achieving a wider variety of ranks. The first and last couple of algorithms nevertheless show less variation in their results and stand out from the other algorithms.

\subsubsection{Visualization of ranking stability}\label{sec:realSingle:stability}

\begin{figure}
    \centering
    \includegraphics[width=1.1\figurewidth]{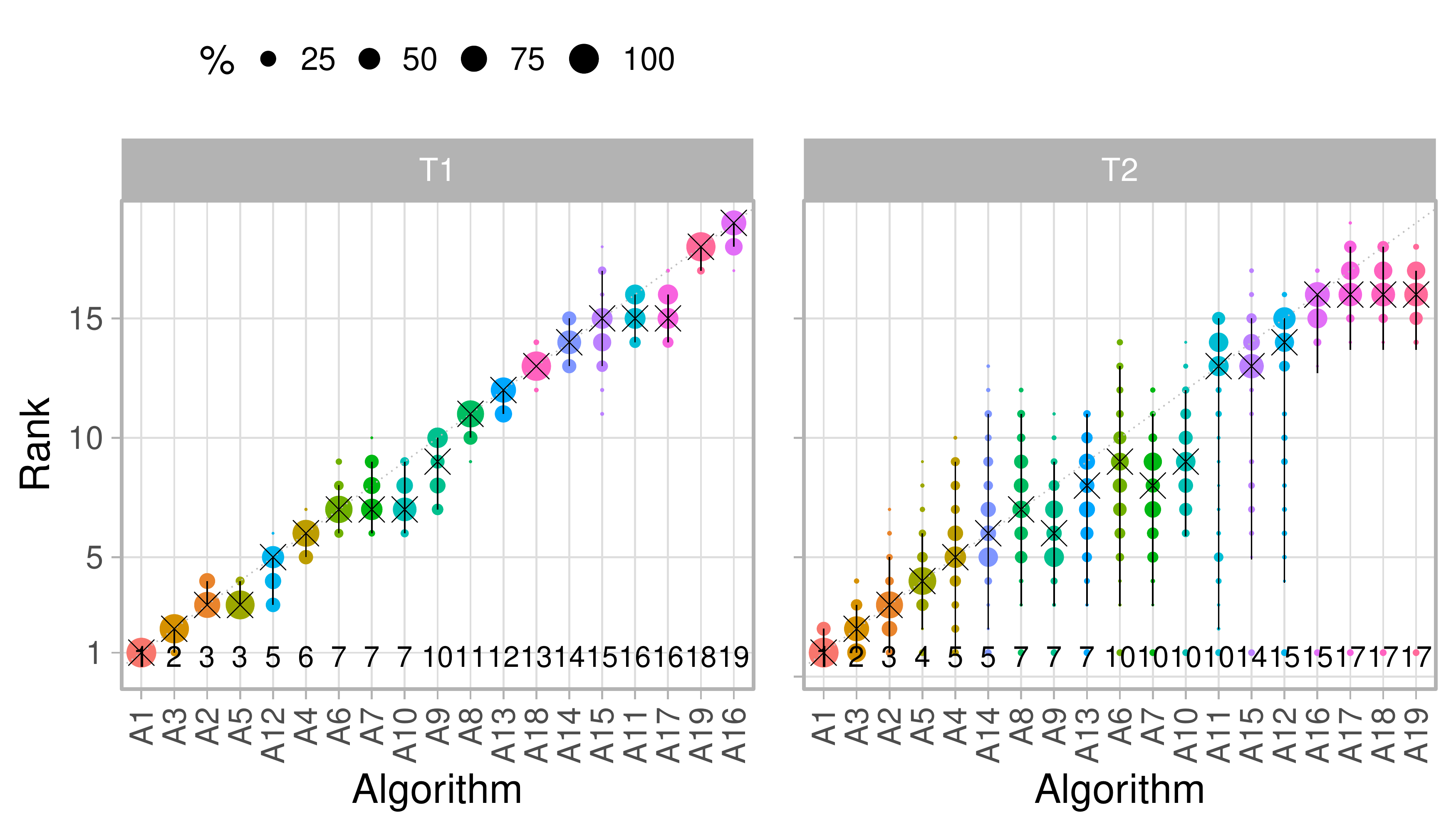}
\caption{Blob plots for selected tasks of the MSD visualize bootstrap results. $T_1$/$T_2$: task with stable/unstable ranking. Ranks above algorithm names highlight the presence of ties.}\label{fig:realSingle:stability:bootstrap}
\end{figure}

The almost diagonal blob plot shown in Fig.~\ref{fig:realSingle:stability:bootstrap} suggests that task $T_1$ leads to relatively clear ranking, whereas $T_2$ shows less stable separation of the algorithms.
In $T_1$, the winning algorithm $A_1$ is ranked first in all bootstrap samples, as is apparent from the fact that no other dot is shown, and the 95\% bootstrap interval consequently only covers the first rank. Only the bootstrap interval of algorithm $A_2$ occasionally covers the first rank (which is thus the winner in some bootstrap samples, together with $A_1$). The rank distributions of all algorithms are quite narrow. In contrast to this relatively clear picture, the blob plot for $T_2$ shows far more ranking variability. Although $A_1$ ranks first for most of the bootstrap samples, the second algorithm also achieves rank 1 in a substantial proportion. 
Most of the algorithms spread over a large range of ranks, for instance the 95\% bootstrap interval for  $A_5$ covers ranks 4 to 13. 
The 
four
last-ranked algorithms separate relatively clearly from the rest. 
Interestingly, all of the algorithms achieved rank 1 in at least one bootstrap sample. This occurred because \textit{significance ranking} produced the same result for all algorithms, which were thus assigned to rank 1 in at least 13 bootstrap samples.
Note that bootstrapping in case of few test cases should be treated with caution since the bootstrap distribution may not be a good estimate of the true underlying distribution. 

The violin 
plots 
shown in Fig.~\ref{fig:realMultiple:task:bootstrap:violin} 
illustrate another perspective on bootstrap sampling. They show the distribution of correlations between rankings based on the full assessment data, and each bootstrap sample in terms of Kendall's $\tau$ for all tasks. A narrow density for high values suggests a stable overall ranking for the task. 
Focusing on tasks $T_1$ and $T_2$, this
again confirms that $T_1$ leads to stable ranking and $T_2$ leads to less stable ranking.



\begin{figure}
\centering
\begin{subfigure}[b]{.55\figurewidth}
    \centering
   \caption{Task $T_1$}\label{subfig:realSingle:stability:significance:a}
    \includegraphics[width=.58\figurewidth]{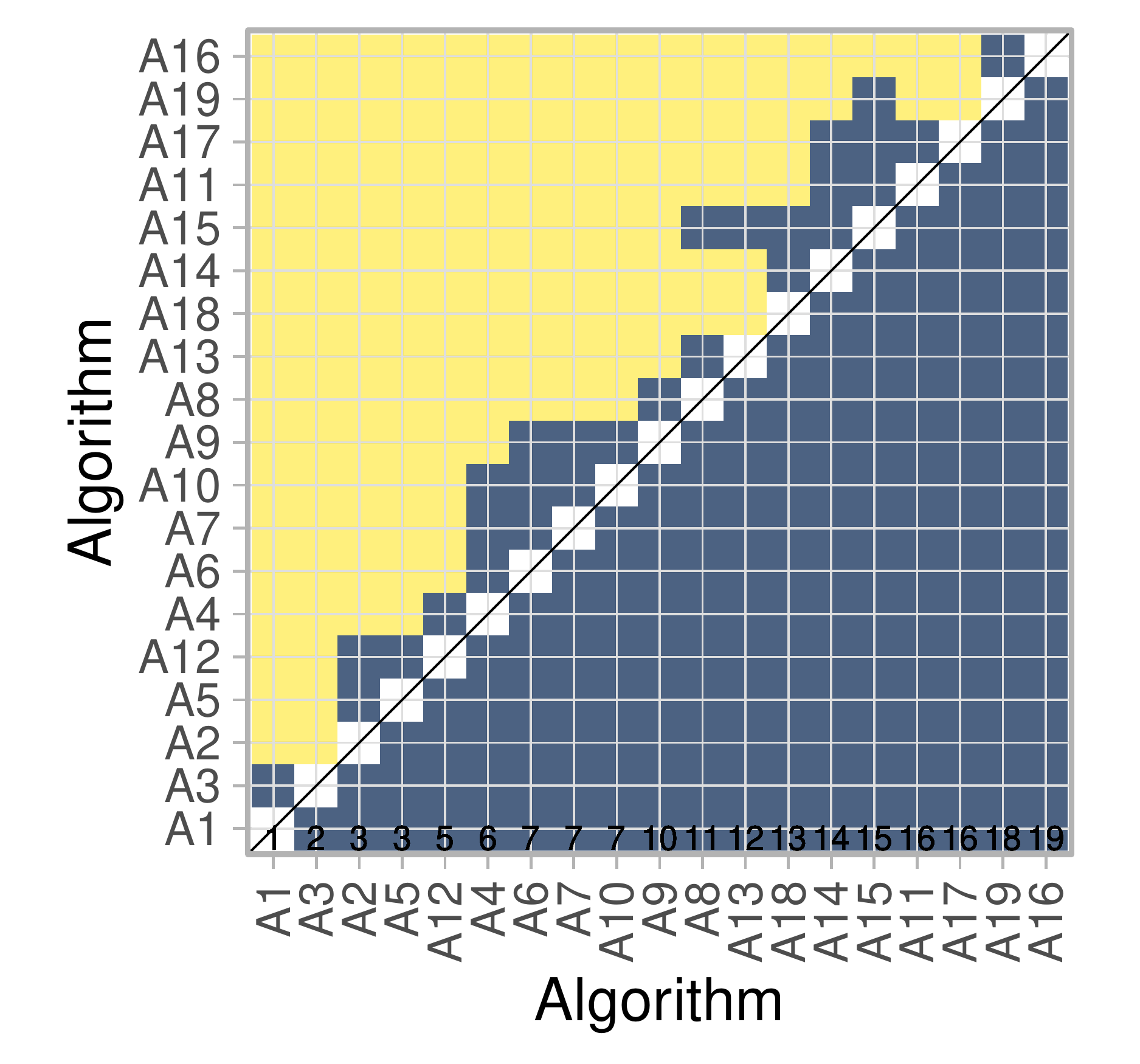}
 \end{subfigure}
\begin{subfigure}[b]{.55\figurewidth}
    \centering
    \caption{Task $T_2$}\label{subfig:realSingle:stability:significance:b}
    \includegraphics[width=.58\figurewidth]{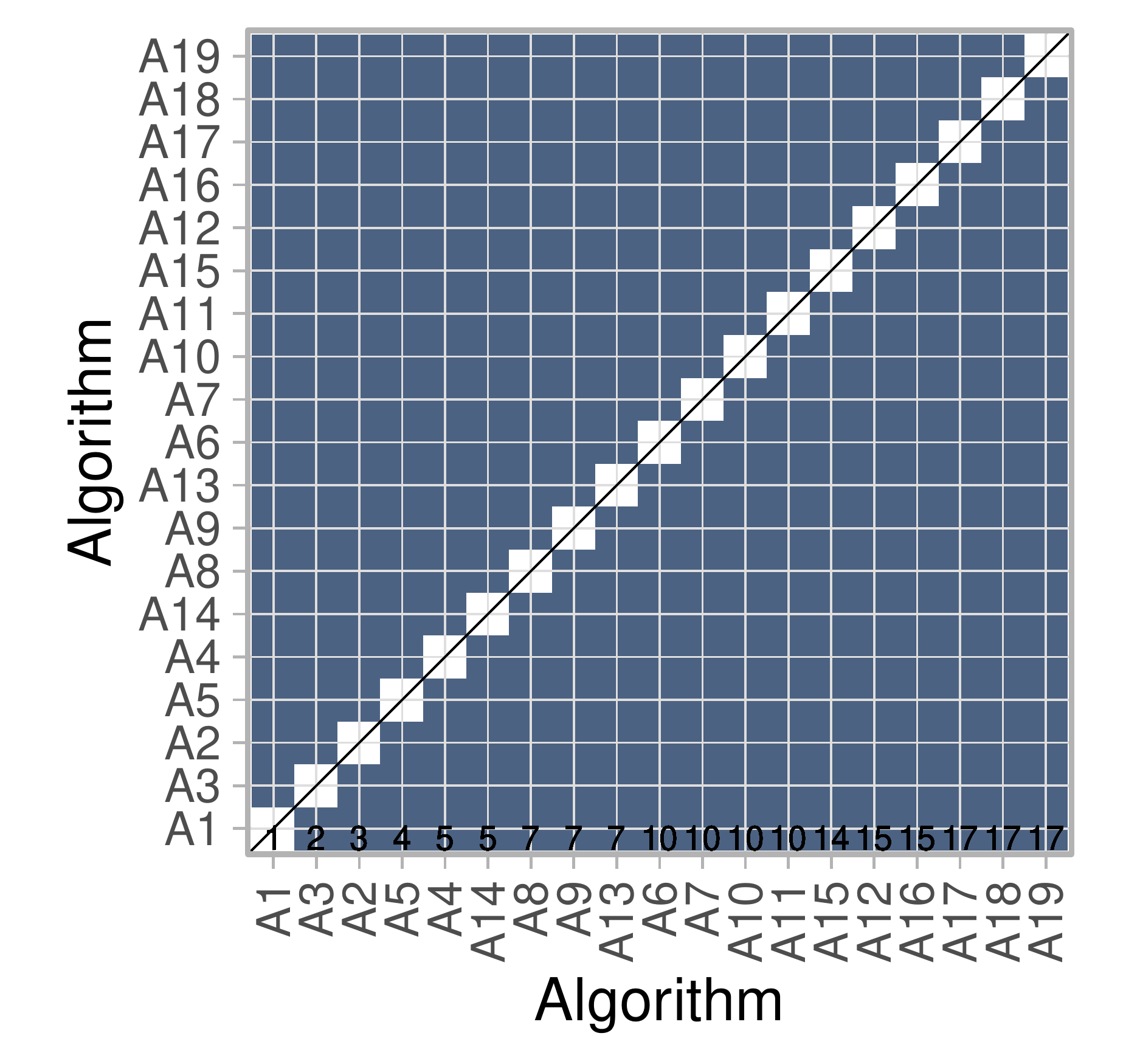}
\end{subfigure}
\caption{Significance maps for selected tasks of the MSD for visualizing the results of significance testing. 
}\label{fig:realSingle:stability:significance}
\end{figure}

The significance map in Fig.~\ref{fig:realSingle:stability:significance} confirms that task $T_1$ provides a clear ranking of the algorithms with the two top ranked algorithms separating from the remaining algorithms, while in $T_2$ the uncertainty is too large to provide a meaningful ranking.


\begin{figure}
\centering
\includegraphics[width=.9\figurewidth]{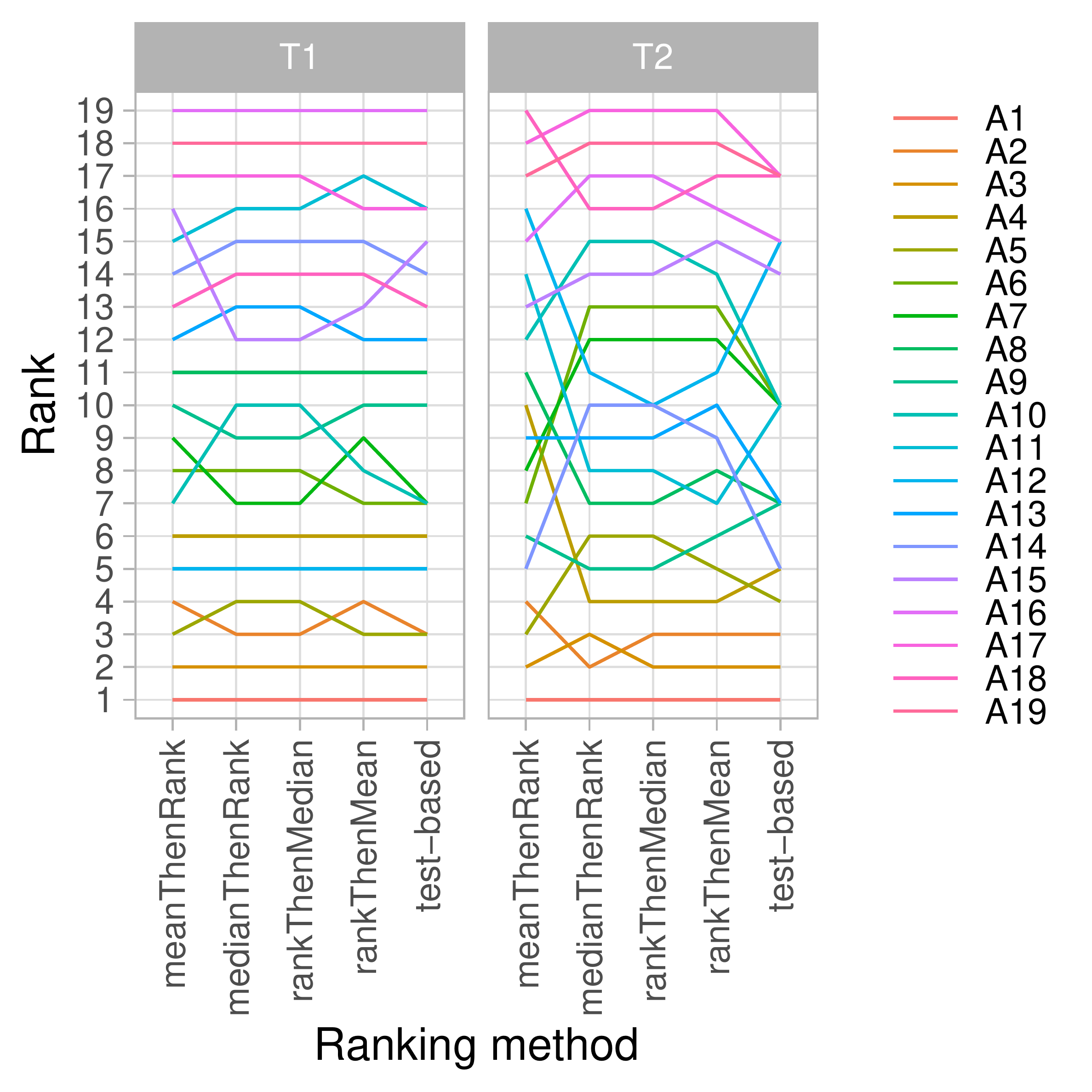} 
\caption{Line  plots for  visualizing  rankings  robustness across  different  ranking  methods. 
}\label{fig:realSingle:stability:method}
\end{figure}

Fig.~\ref{fig:realSingle:stability:method} depicts ranking lists from different methods, confirming that in $T_1$, rankings are relatively robust across ranking methods. Rankings in $T_2$ depend far more on the ranking method. Furthermore, many algorithms attain the same rank in the test-based procedure, a pattern which is often observed in challenges with unclear ranking. Interestingly, ranking according to average DSC (\textit{mean-then-rank}) leads to a considerably different ranking than (nonparametric) test-based ranking, suggesting that the outlying test cases mentioned in sec.~\ref{sec:realSingle:raw} have a strong impact on the former ranking.

\subsection{Visualization of cross-task insights}\label{sec:realMultiple}
All nine tasks in the real world assessment data set were used as an example for multi-task analyses. 

As previously mentioned, an aggregation (consensus) of rankings across tasks is needed to order the algorithms along the x-axes or in panels. For the present example, we have taken the average rank after \textit{significance ranking} on a task basis (see \ref{sec:realSingle}) as consensus. 

\subsubsection{Characterization of algorithms} \label{sec:realMultiple:stability}
\begin{figure}
\centering
\includegraphics[width=\figurewidth]{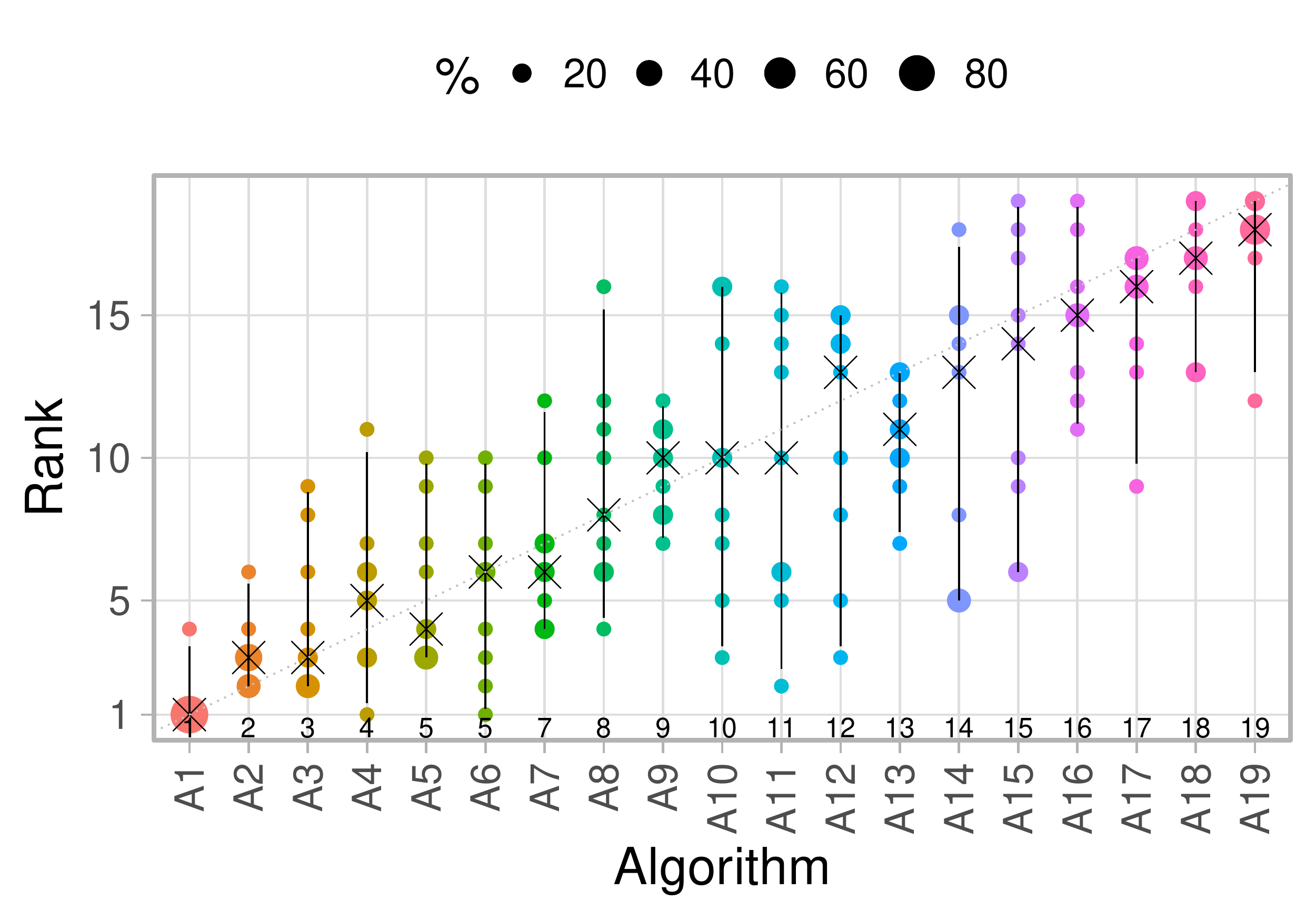} 
\caption{Blob plots for visualizing ranking stability across tasks. Consensus rankings above algorithm names highlight the presence of ties.}\label{fig:realMultiple:stability:raw}
\end{figure}

The first visualization of stability of rankings across tasks is provided in Fig.~\ref{fig:realMultiple:stability:raw}. 
The plot illustrates that $A_1$ almost always ranks first across tasks and only ranks third a few times. The other algorithms achieve a large range of ranks across tasks, apart from the last ranked algorithms, which perform unfavorably in most tasks. 

\begin{figure}
\centering
\begin{subfigure}[b]{1\figurewidth}
    \centering
    \caption{Stratified by algorithm}
    \label{subfig:realMultiple:stability:bootstrap:a}
\includegraphics[width=1.1\figurewidth]{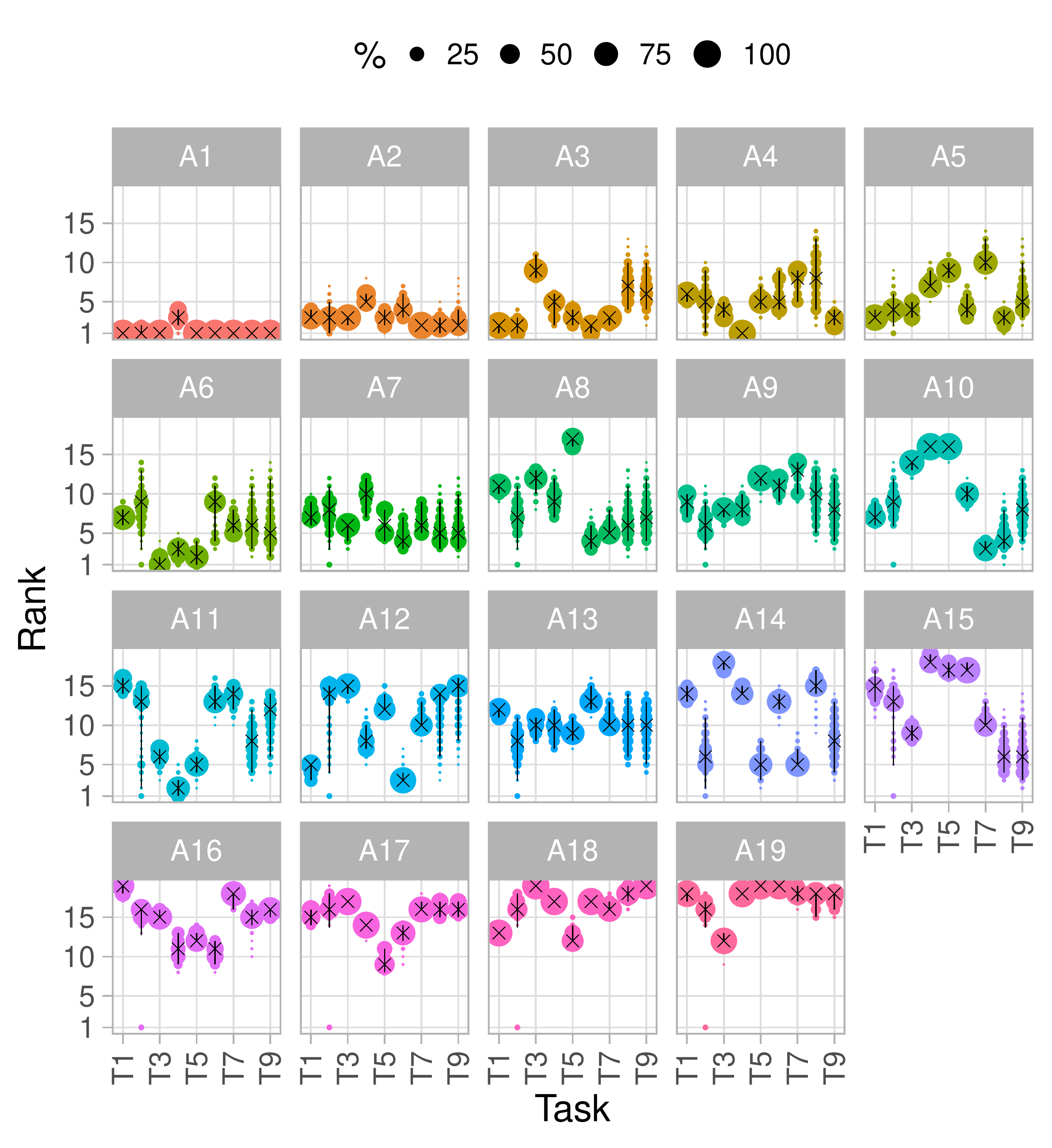} 
\end{subfigure}
\begin{subfigure}[b]{1\figurewidth}
    \centering
    \caption{Stratified by task}
    \label{subfig:realMultiple:stability:bootstrap:b}
\includegraphics[width=1.1\figurewidth]{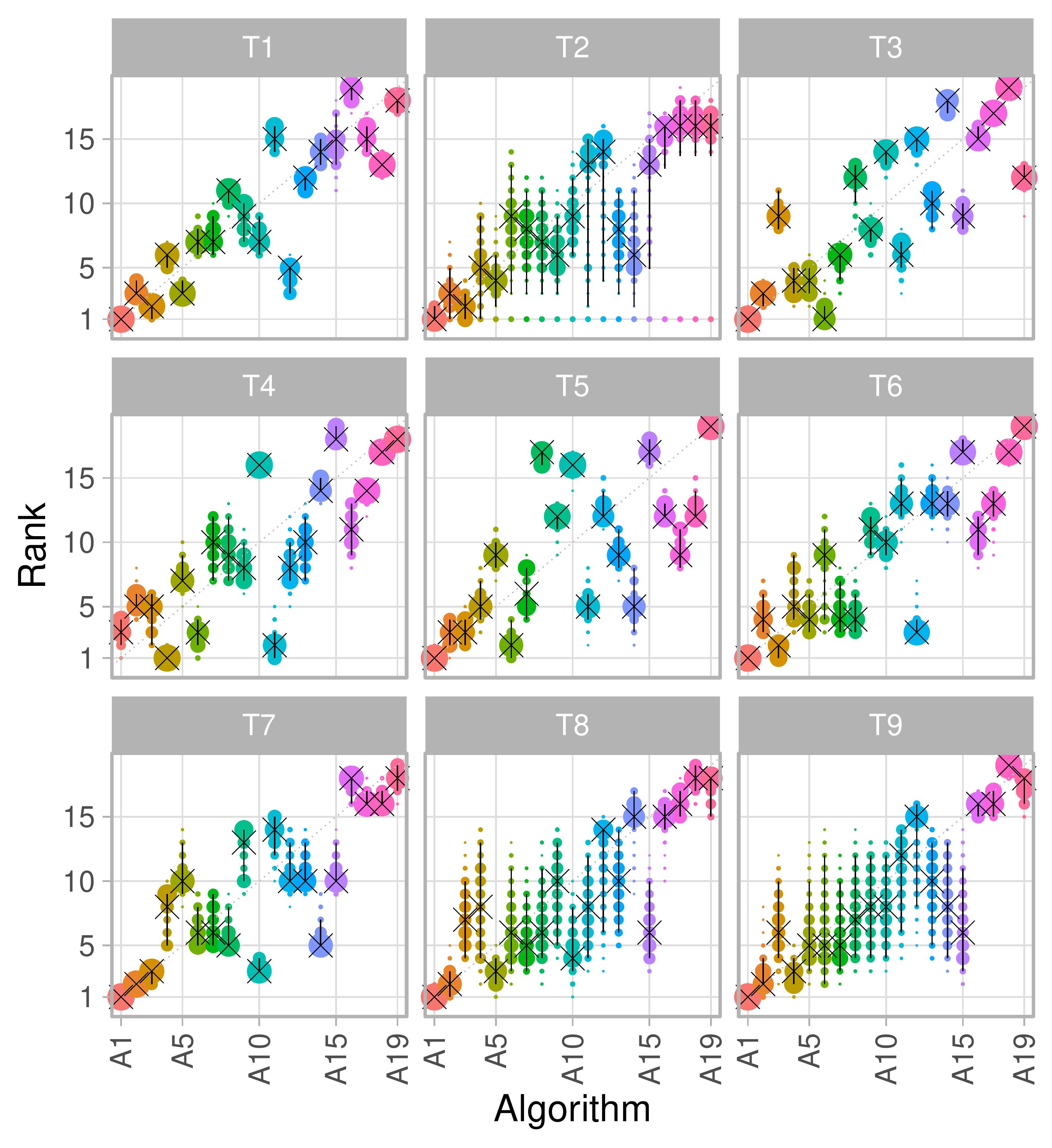} 
\end{subfigure}
\caption{Rank distributions of each algorithm across bootstrap samples stratified by algorithm (top) and task (bottom).}\label{fig:realMultiple:stability:bootstrap}
\end{figure}

The blob plot of bootstrap results across tasks (Fig.~\ref{subfig:realMultiple:stability:bootstrap:a}) gives detailed insights into the performance of each algorithm.
The first ranked algorithm ($A_1$) is almost always among the winners in each task, and only task $T_4$ stands out; as such, it is very stable. $A_1$ never attains a rank worse than four. Although the second-ranked algorithm ($A_2$) performs worse than $A_1$, it consistently attains top ranks as well, apart from $T_4$. 
Despite 
$A_3$, $A_4$ and $A_5$
being among the winners in some tasks, they show vastly variable metric values across tasks. Medium-ranked algorithms are either in the midrange in all tasks (e.g., $A_9$), 
or perform reasonably well in a few tasks and fail in others (e.g., $A_{10}$).

%
%

\subsubsection{Characterization of tasks} \label{sec:realMultiple:task}

To visualize which tasks separate algorithms well (i.e., lead to a stable ranking), we have rearranged the data from Fig.~\ref{subfig:realMultiple:stability:bootstrap:a} and have shown the bootstrap results for all algorithms separately by task, see Fig.~\ref{subfig:realMultiple:stability:bootstrap:b}. From this plot, we can see that task $T_1$ apparently leads to stable rankings (but not necessarily on the diagonal, i.e., different from the consensus ranking), whereas rankings from tasks $T_2$ and $T_9$ are far more variable, or at least this is the case for medium-ranked algorithms. 

\begin{figure}
\centering
\includegraphics[width=\figurewidth]{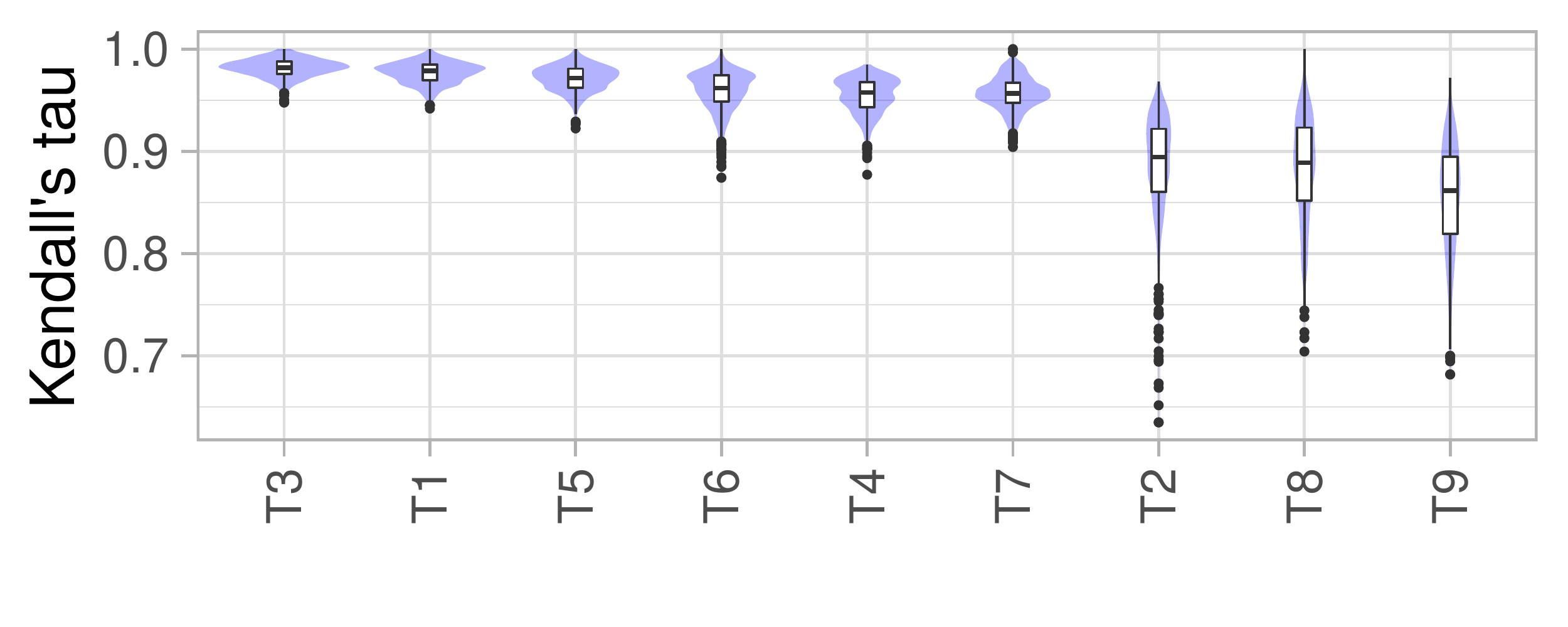} 
\caption{Violin plots for comparing ranking stability across tasks arranged by median Kendall's $\tau$.}\label{fig:realMultiple:task:bootstrap:violin}
\end{figure}

Another view of the bootstrap results is provided by violin 
plots (see Fig.~\ref{fig:realMultiple:task:bootstrap:violin}), which show the distribution of Kendall's $\tau$ between the ranking based on the full assessment data set and the ranking for each bootstrap sample. Tasks $T_1$, $T_3$ and $T_5$ provide very stable rankings for all algorithms; $T_4$, $T_6$ and $T_7$ are slightly less stable overall because a subset of algorithms does not separate well. $T_2$, $T_8$ and $T_9$ yield the least stable ranking overall.

%
%

The similarity/clustering of tasks with respect to their algorithm rankings is visualized in a dendrogram and network-type graph in Fig.~\ref{fig:realMultiple:task:clustering}. In both cases, Spearman's footrule distance is used and complete agglomeration is applied for the dendrogram. Distances between nodes are chosen to increase exponentially in Spearman's footrule distance with a growth rate of 0.05 to accentuate large distances. 

While the dendrogram suggests two major clusters of tasks, the network-type graph highlights that $T_5$ in particular seems to be different from the remaining tasks in terms of its ranking. It also highlights $A_1$ as the winner in most tasks.

\section{Discussion}
\label{sec:conclusion}

While the significance of biomedical challenges is growing at an enormous pace, the topic of analysis and visualization of assessment data has received almost no attention in the literature to date. In this context, the contributions of this paper can be summarized as follows: 
\begin{enumerate}
    \item Methodology
    : To our knowledge, we are the first to propose a systematic way to analyze and visualize the results of challenges in general and of multi-task challenges in particular.
    \item Open source visualization toolkit (challengeR \cite{challengeR}): The methodology was implemented as an open-source R \cite{R2019} toolkit to enable quick and wide adoption by the scientific community. 
    \item Comprehensive validation: The toolkit was applied to a variety of simulated and real challenges. According to our results, it offers an intuitive way to extract important insights into the performance of algorithms, which cannot be revealed by commonly applied presentation techniques such as ranking tables and boxplots. 
\end{enumerate}

While the assessment of uncertainty in results is common in many fields of quantitative analysis, it is surprising that uncertainty in rankings in challenges has seemingly been neglected. To address this important topic, this work places particular focus on the analysis and visualization of uncertainties. 

It should be noted that visualization methods often reach their limit when the number of algorithms is too large. In this case, data analysis can be performed on all algorithms, but visualization can be reduced to a top list of algorithms.

Whereas the methodology and toolkit proposed were designed specifically for the analysis and visualization of challenge data, they may also be applied to presenting the results of validation studies performed in the scope of classical original papers. In these papers it has become increasingly common to compare a new methodological contribution with other previously proposed methods. Our methods can be applied to this use case in a straightforward manner. Similarly, the toolkit has originally been designed for the field of biomedical image analysis but can be readily applied in many other fields.

In conclusion, we believe that our contribution could become a valuable tool for analyzing and visualizing challenge results. Due to its generic design, its impact may reach beyond the field of biomedical image analysis. 

\section*{Acknowledgements}
This project was conducted in the scope of the Helmholtz Imaging Platform (HIP) funded by the Helmholtz Association of German Research Centres. We thank Dr. Jorge Bernal for constructive comments on an earlier version.




\ifCLASSOPTIONcaptionsoff
  \newpage
\fi

\bibliography{bibliographies/main.bib}
\bibliographystyle{unsrt}

\end{document}